\documentclass[
  onecolumn,
  natbib,
]{miri-tech-article}
\usepackage[
  notodo,
  nocitefix,
]{miritools}
\usepackage[T1]{fontenc}
\usepackage{lmodern}
\usepackage{float}
\usepackage{bm}
\usepackage{graphicx}
\usepackage{enumerate}
\usepackage[noabbrev]{cleveref}
\usepackage[font=small,labelfont=bf]{caption}
\usepackage[hang,flushmargin]{footmisc}
\usepackage{setspace}

\usepackage{tocloft}

\setcitestyle{numbers,square}
\begin{document}

\title{An overview of 11 proposals for building safe advanced AI}
\author{Evan Hubinger\thanks{Research supported by the Machine Intelligence Research Institute (intelligence.org). Special thanks to Kate Woolverton, Paul Christiano, Rohin Shah, Alex Turner, William Saunders, Beth Barnes, Abram Demski, Scott Garrabrant, Sam Eisenstat, and Tsvi Benson-Tilsen for providing helpful comments and feedback. Additional thanks to Claire Wang and Rob Bensinger for helping create this paper out of the Alignment Forum post that preceded it.\cite{post}} \\ Research Fellow, Machine Intelligence Research Institute}

\date{May 29, 2020}

\maketitle

\begin{abstract}
This paper analyzes and compares 11 different proposals for building safe advanced AI under the current machine learning paradigm, including major contenders such as iterated amplification, AI safety via debate, and recursive reward modeling. Each proposal is evaluated on the four components of outer alignment, inner alignment, training competitiveness, and performance competitiveness, of which the distinction between the latter two is introduced in this paper. While prior literature has primarily focused on analyzing individual proposals, or primarily focused on outer alignment at the expense of inner alignment, this analysis seeks to take a comparative look at a wide range of proposals including a comparative analysis across all four previously mentioned components.
\end{abstract}

\tableofcontents

\section{Introduction}
\label{sec:0}

This paper is a collection of 11 different proposals for building safe advanced AI under the current machine learning paradigm. While there already exist good overviews of various particular approaches, such as amplification\cite{amplification}, debate\cite{debate}, or recursive reward modeling\cite{leike}, most of that literature focuses primarily on outer alignment at the expense of inner alignment\cite{risks} and doesn't provide direct comparisons between approaches. The goal of this paper is to address this problem by providing a single collection of 11 different proposals for building safe advanced AI---each including both inner and outer alignment components.\footnote{Note that the order of the proposals given here is chosen purely for pedagogical reasons and is not meant to imply any sort of measure of importance.}\textsuperscript{,}\footnote{For another proposal evaluated in the same way as those presented here that came out after this list was completed in May 2020, see ``AI safety via market making.''\cite{market_making}}

This paper does not cover all existing proposals, and I strongly expect that there will be many additional new proposals in the future. Nevertheless, I think it is quite useful to take a broad look at what we have now and compare and contrast some of the current leading candidates.

It is important to note that the way I describe the 11 approaches presented here is not necessarily an accurate representation of how others would describe them. You should treat the approaches I describe here as a particular version of the approach that I selected for analysis, rather than as a canonical version that their various creators or proponents would endorse.

Furthermore, this paper only includes approaches that intend to directly build advanced AI systems via machine learning. Thus, this paper does not include other possible approaches for solving the broader AI existential risk problem such as:
\begin{itemize}
\item finding a fundamentally different way of approaching AI than the current machine learning paradigm, with the intent of starting from foundations that make it easier to build safe advanced AI,
\item developing some advanced technology that produces a decisive strategic advantage\cite{superintelligence} without using advanced AI, or
\item achieving global coordination around not building advanced AI via (for example) a persuasive demonstration that any advanced AI is likely to be unsafe.
\end{itemize}
For each of the proposals that I consider, I will try to evaluate them on the following four basic components that I think are necessary in any story about how to build safe advanced AI under the current machine learning paradigm.

\begin{enumerate}
\item \textbf{Outer alignment} -- Outer alignment is about asking why the objective we're training for is aligned---that is, if we actually got a model that was trying to optimize for the given loss/reward/etc., would we like that model? For a more thorough description of what I mean by outer alignment, see ``Outer alignment and imitative amplification.''\cite{outer_alignment}
\item \textbf{Inner alignment} -- Inner alignment is about asking the question of how our training procedure can guarantee that the model it produces will, in fact, be trying to accomplish the objective we trained it on. For a more rigorous treatment of this question and an explanation of why it might be a concern, see ``Risks from Learned Optimization in Advanced Machine Learning Systems.''\cite{risks} We'll be assuming familiarity with concepts from this paper, such as \emph{mesa-optimization}, \emph{deceptive alignment}, and \emph{pseudo-alignment}, below.
\item \textbf{Training competitiveness} -- Competitiveness is a bit of a murky concept, so I've broken it up into two pieces here. Training competitiveness is the question of whether the given training procedure is one that a team or group of teams with a reasonable lead would be able to afford to implement without completely throwing away that lead. Thus, training competitiveness is about whether the proposed process of producing advanced AI is competitive.
\item \textbf{Performance competitiveness} -- Performance competitiveness, on the other hand, is about whether the final product produced by the proposed process is competitive. Performance competitiveness is thus about asking whether a particular proposal, if successful, would satisfy the use cases for advanced AI---e.g. whether it would fill the economic niches that people want artificial general intelligence (AGI) to fill.
\end{enumerate}

\noindent It can be easy to focus too much on either the alignment side or the competitiveness side, while neglecting the other. We obviously want to avoid proposals which could be unsafe, but the ``do nothing'' proposal is equally unacceptable---while doing nothing is quite safe in terms of having no chance of directly leading to existential risk, it doesn't actually help in any way relative to what would have happened by default. We therefore want proposals that are both aligned and competitive: not only should the resulting system not result in catastrophe, but it should also help substantially reduce existential risk in general---by providing a model of how safe advanced AI can be built, by being powerful enough to assist with future alignment research, and/or by granting a decisive strategic advantage that can be leveraged into otherwise reducing existential risk.

\section{Reinforcement learning + transparency tools}
\label{sec:1}

We want to be able to train advanced AI systems to learn what we want; and to the extent the system doesn't fully understand what we want, we want the system to be \emph{corrigible}, tolerating (or even assisting in) corrections, safeguards, and improvements.\cite{corrigibility}\cite{mechanistic}

The first proposal we'll consider tries to achieve this via reinforcement learning (RL) and transparency tools:

\begin{enumerate}
\item Train an RL agent in an environment where corrigibility, honesty, multi-agent cooperation, etc. are incentivized. The basic idea would be to mimic the evolutionary forces that led to humans' general cooperativeness. As an example of work along these lines that exists now, see OpenAI's hide and seek game.\cite{tool_use} The environment could additionally be modified to directly reward following human instructions, thus encouraging corrigibility towards humans. For a more thorough discussion of this possibility, see Richard Ngo's ``Multi-agent safety.''\cite{multi_agent_safety}

\vspace{4mm}

\begin{figure}[h!]
  \centering
  \includegraphics[width=0.7\textwidth]{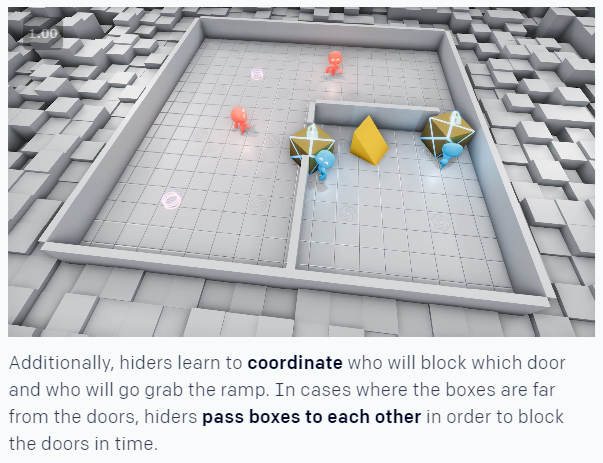}
  \caption{An image of OpenAI's hide and seek game.}
\end{figure}

\vspace{2mm}

\item Have humans use transparency tools,\cite{circuits} adversarial training, etc. to check for deceptive\cite{risks} or otherwise catastrophic\cite{catastrophes} behavior in the resulting model.
\end{enumerate}

\subsection{Outer alignment} Whether this proposal satisfies outer alignment is entirely dependent on whatever the dominant behavior is in the training environment---that is, it depends on the deployment behavior of those models which perform optimally in the training environment. If corrigibility, honesty, cooperation, etc. do in fact dominate in the limit, then such an approach would be outer aligned. 

By default, however, it seems quite difficult to understand the limiting behavior of complex, multi-agent environments, especially if they are anywhere as complex as the actual human ancestral environment. If following human instructions is incentivized, for example, that could lead to corrigibility in the limit---or it could lead to agents that only choose to follow human instructions for the instrumental reason of believing it will help them acquire more resources. 

Alternatively, it might be possible to isolate the structure that was present in the human ancestral environment that led us to be cooperative, honest, etc. One worry here, however, is that even if we could figure out how to properly incentivize cooperation, it might result in agents that are cooperative with each other but not very cooperative with us, similar to how we might not be very cooperative with aliens that are very different from us.

\subsection{Inner alignment} Inner alignment in this setting requires that the system's training produces something in line with the optimal behavior in the environment rather than other, potentially perverse equilibria---the alignment of this optimal behavior with our actual preferences is then an outer alignment question.

The basic proposal for how to avoid such perverse equilibria under this proposal is via the use of checks such as transparency tools and adversarial training to detect inner alignment failures before the model is deployed. Chris Olah describes this sort of transparency checking as giving you a ``mulligan'' that lets you throw away your model and start over if you find something wrong.\cite{chris_olah} Thus, ideally, if this approach ends up not working it should be clear before the model is deployed, allowing either for this approach to be fixed or for a new approach to be found instead. There is also a reasonable chance that it does work initially---we don't understand our models' inductive biases very well, but it seems entirely possible that they could work out such that pseudo-alignment\cite{risks} is disincentivized.

While it seems quite plausible to me that this sort of approach could catch proxy pseudo-alignment, it seems unlikely to me that it would successfully catch deceptive pseudo-alignment, as it could be very difficult to make transparency tools that are robust to a deceptive model actively trying to trick them. To catch deceptive alignment, it seems likely to be necessary to incorporate such checks into the training process itself---which is possible to do in this setting, though is not the approach I described above---in order to prevent deception from occurring in the first place, rather than trying to detect it after the fact.

\subsection{Training competitiveness} Training competitiveness here seems likely to depend on the extent to which the sort of agency produced by RL is necessary to train advanced AI systems.

Performing RL in highly complex, difficult-to-simulate environments---especially if those environments involve interaction with the real world---could be quite expensive from a training competitiveness standpoint. Compared to simple language modeling, for example, the difficulty of on-policy data collection combined with low sample-efficiency could make full-scale RL much less training competitive. These sorts of competitiveness concerns could be particularly pronounced if the features necessary to ensure that the RL environment is aligned result in its being significantly more difficult to simulate.

That being said, if RL is necessary to do anything powerful and simple language modeling is insufficient, then whether or not language modeling is easier is a moot point. Whether RL is really necessary seems likely to depend on the extent to which it is necessary to explicitly train agents---which is very much an open question. Additionally, even if agency is required, it could potentially be obtained just by imitating an actor such as a human that already has it, rather than training it directly via RL.

\subsection{Performance competitiveness} The question for performance competitiveness in this setting is to what extent it is possible to create an environment that incentivizes all of the behavior you might want from your AGI. Such an environment doesn't need to be purely simulated---you could do some simulation training and some real-world training, for example. Regardless of how your RL environment is constructed, however, it needs to actually incentivize the correct behavior for the tasks that you want to use your AI for.

For example: can you incentivize good decision-making? Good question-answering? Good learning ability? Do you need good fine motor control, and if so, can you incentivize it?

These are highly non-trivial questions: it could be quite difficult to set up an RL environment to teach an agent to do all of the tasks you might want it to perform to fill all the economic niches for AGI, for example. This will of course be highly dependent on exactly what economic niches you want your advanced AI to fill.

\section{Imitative amplification + intermittent oversight}
\label{sec:2}

Though many of the approaches on this list make use of the basic iterated amplification\cite{amplification} framework, imitative amplification is probably the most straightforward. Still, imitative amplification has a number of moving parts.

To define imitative amplification, we'll first define $\text{Amp}(M)$---the ``amplification operator''---as the procedure where a human $H$ answers a question with access to a model $M$.\footnote{Note that in practice $\text{Amp}(M)$ doesn't need to always be computed with an actual human. $H$ can simply be $M$ some fixed fraction of the time, for example---or more often when $M$ is confident and less often when $M$ is uncertain---and the resulting procedure is effectively the same. See ``A concrete proposal for adversarial IDA''\cite{adversarial_ida} for an example of how something like that could work.}

\vspace{4mm}

\begin{figure}[H]
  \centering
  \includegraphics[width=\textwidth]{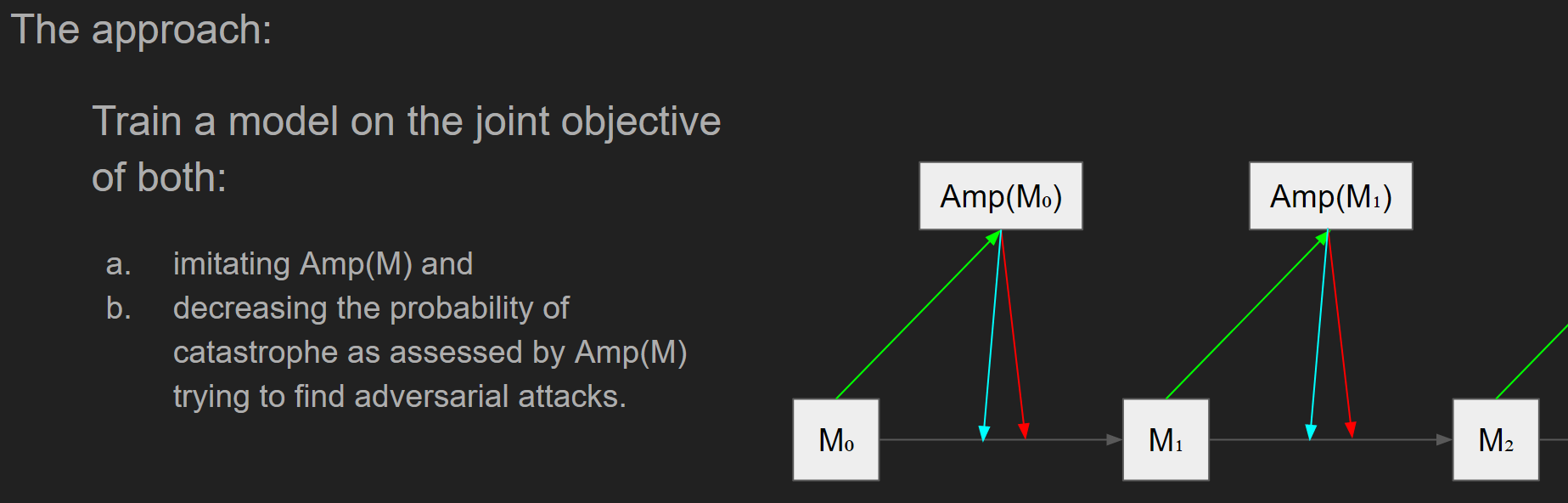}
  \caption{A diagram of the amplification operator $\text{Amp}(M)$, where white arrows indicate information transfer, $Q$ is a question, $A$ is $\text{Amp}(M)$'s answer, $H$ is a human, and $M$ is the model.}
\end{figure}

\noindent Then, imitative amplification is just the procedure of iteratively training $M$ to imitate $\text{Amp}(M)$.

\vspace{4mm}

\begin{figure}[H]
  \centering
  \includegraphics[width=\textwidth]{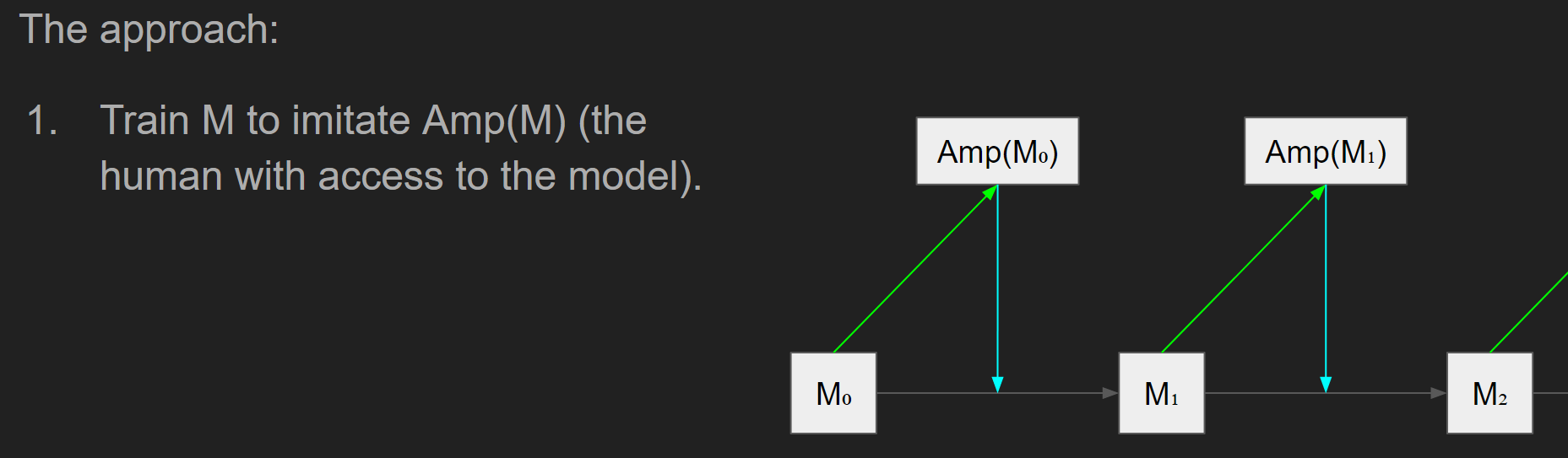}
  \caption{The basic imitative amplification setup, where green arrows indicate amplification, gray arrows indicate training, and cyan arrows indicate the imitative amplification loss.}
\end{figure}

\noindent Finally, we can define imitative amplification + intermittent oversight---which is the full approach we want to consider here---as the combination of imitative amplification with intermittent oversight of $M$ by $\text{Amp}(M)$ whenever the target model changes. Specifically, we want $\text{Amp}(M)$ to look for deceptive or otherwise catastrophic behavior in $M$ by utilizing things like transparency tools and adversarial attacks.

\vspace{4mm}

\begin{figure}[H]
  \centering
  \includegraphics[width=\textwidth]{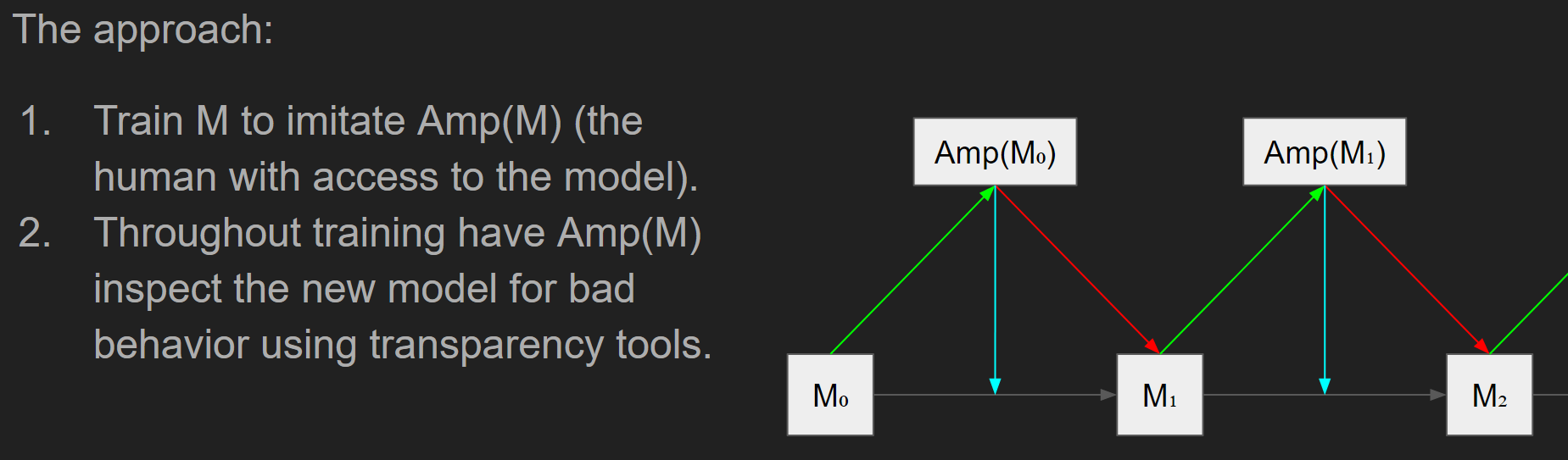}
  \caption{Imitative amplification plus intermittent oversight, where red arrows indicate oversight (from the overseer to the overseen model).}
\end{figure}

\vspace{2mm}

\subsection{Outer alignment} Since imitative amplification trains $M$ to imitate $\text{Amp}(M)$, it limits to the fixed point of the $\text{Amp}$ operator, which Paul Christiano calls HCH for ``Humans Consulting HCH.''\footnote{Imitative amplification leads to HCH in the limit of training a perfect predictor over all possible inputs---that is, an infinite amount of training data, an infinitely large model, and an infinitely large amount of training time.}\cite{strong_hch} HCH is effectively a massive tree of humans consulting each other to answer questions.

\vspace{4mm}

\begin{figure}[h!]
  \centering
  \includegraphics[width=\textwidth]{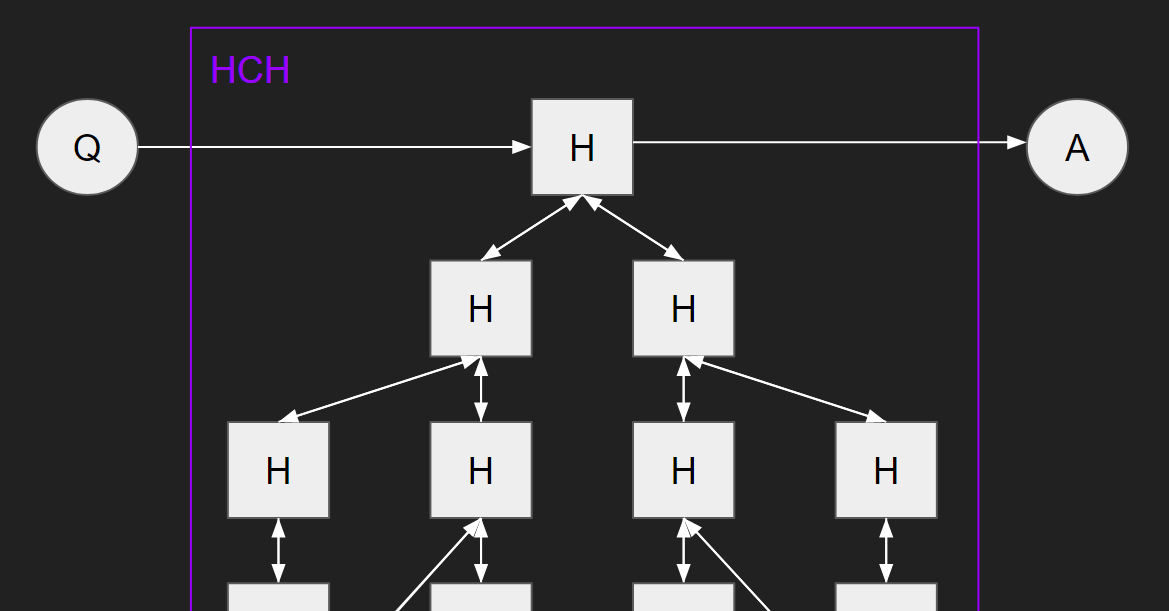}
  \caption{A (partial) diagram of HCH, where white arrows indicate information transfer.}
\end{figure}

\vspace{2mm}

\noindent Thus, whether imitative amplification is outer aligned is dependent on whether HCH is aligned or not. HCH's alignment, in turn, is likely to depend heavily on the specific humans used and what sort of policy they are implementing. The basic idea, however, is that since the limit is composed entirely of humans---and since we can control what those humans do---we should be able to ensure that all the optimization power is coming from the humans (and not from memetic selection, for example), which hopefully should allow us to make it safe. While there are certainly valid concerns with the humans in the tree accidentally (or purposefully) implementing some misaligned policy, there are possible things that can be done to address these problems.\cite{universality}

\subsection{Inner alignment} The hope here is to be able to detect and fix serious inner alignment failures via the use of intermittent oversight---that is, inspection by a human with access to a previous version of the model. Thus, the question of inner alignment here is dependent on the question of whether such an overseer would be able to detect deceptive or otherwise catastrophic behavior in the model.

Ideally, if such detection is working, it should be clear as soon as the model starts to become misaligned, giving us the opportunity to either fix the issue or find a new approach. For that to work, however, it will likely need to be the case that the intermittent oversight is performed regularly enough to catch any deception as soon as it starts appearing---since once a model becomes deceptive, it might start trying to obfuscate its deception, which would make it significantly harder for the overseer to detect problems.

That being said, while it might be quite difficult to \textit{directly} detect deceptive cognition early enough, it might be possible to instead detect deceptive behavior \textit{indirectly} by verifying a condition that rules out deception while still being relatively easy to verify---such as myopia. In this context, a myopic model would be one that determines its output based on some fixed standard such as distance to HCH while a non-myopic model would be one that determines its output based on its consequences.\footnote{Enforcing myopia for amplification could also have the side benefit of improving competitiveness, as a non-myopic amplification model might try to make its output simpler---and thus less useful---if it is trying to make a human consulting it easier to predict.} For a more detailed explanation of how such myopia verification might work, see ``Towards a mechanistic understanding of corrigibility.''\cite{mechanistic}

Additionally, because we do intermittent oversight whenever we change the target network, we have a guarantee that the overseer is always at least as strong as any targets that the model was trained to imitate. That being said, the model could end up stronger than the targets it was trained to imitate if it manages to, for example, achieve significant compression of what the targets are doing---which in fact is necessary for training competitiveness. Furthermore, even if the overseer is stronger than the model, the offense-defense balance could be such that a deceptive model would still be able to hide its deception. Which side is favored in the offense-defense balance is likely to depend heavily on what sorts of transparency tools the overseer has access to.

\subsection{Training competitiveness} First, the training competitiveness of imitative amplification is likely to depend on whether pure imitation can be turned into a rich enough reward signal to facilitate highly sample-efficient learning. In my opinion, it seems likely that human language imitation (where language includes embedded images, videos, etc.) combined with techniques to improve sample efficiency\footnote{See Paul Christiano's ``Efficient feedback''\cite{efficient_feedback} for some examples of possible sample-efficiency-improving techniques.} will be competitive at some tasks---namely highly-cognitive tasks such as general-purpose decision-making---but not at others, such as fine motor control. If that's true, then as long as the primary economic use cases for AGI fall into the highly-cognitive category, imitative amplification should be training competitive. For a more detailed analysis of this question, see ``Outer alignment and imitative amplification.''\cite{outer_alignment}

Second, imitative amplification + intermittent oversight could also fail to be training competitive if it is very difficult for the intermittent oversight check to ever pass. This could happen if, for example, the training process for imitative amplification doesn't naturally produce models that are transparent enough to facilitate such a check. That being said, what counts as ``transparent enough'' again depends heavily on the specifics of the transparency tools that the overseer has access to.

\subsection{Performance competitiveness} The performance competitiveness question here, as with the outer alignment question, is heavily dependent on the properties of HCH, as imitative amplification is specifically designed to limit toward HCH. In particular, performance competitiveness is dependent on whether HCH is sufficiently universal so as to be capable of solving whatever hard problems we might want advanced AI to solve. For example, if trees of humans can't replicate intuitive flashes of insight, and such flashes of insight are necessary to solve the sorts of highly-cognitive tasks we want AGI for, then imitative amplification would fail to be performance competitive. Alternatively, as long as it is always possible for a human to break down complex tasks that they can't solve themselves into easier subtasks, imitative amplification should be performance competitive. Note that this doesn't have to be the case for an arbitrary human without any training---only for the sort of human from whom you would actually be collecting training data.

\section{Imitative amplification + relaxed adversarial training}
\label{sec:3}

The basic idea here is very similar to the previous approach, in that we start out with pure imitative amplification.

\vspace{4mm}
\begin{figure}[h!]
  \centering
  \includegraphics[width=\textwidth]{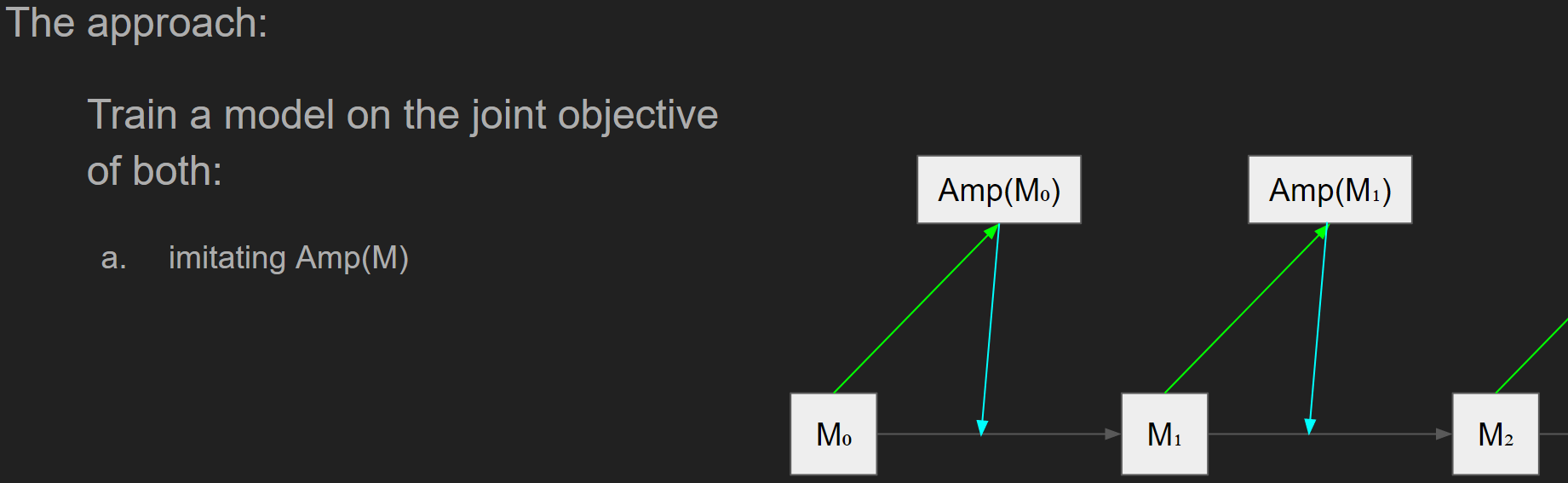}
  \caption{The basic imitative amplification setup.}
\end{figure}
\vspace{2mm}

\noindent However, instead of doing intermittent oversight checks, we train directly on the intermittent oversight signal---that is, we jointly train $M$ on the imitation loss and on decreasing the probability of catastrophe as assessed by $\text{Amp}(M)$---yielding a relaxed adversarial training setup.\cite{relaxed}

\newpage

\vspace{4mm}
\begin{figure}[h!]
  \centering
  \includegraphics[width=\textwidth]{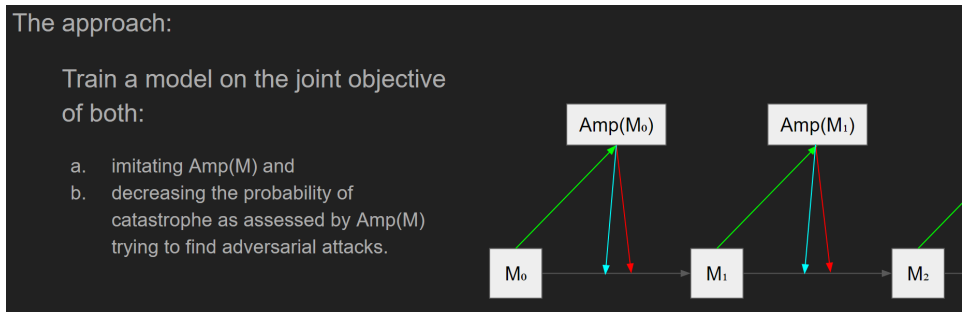}
  \caption{Imitative amplification plus relaxed adversarial training, where the red arrows indicate the oversight loss.}
\end{figure}
\vspace{2mm}

\subsection{Outer alignment} Imitative amplification + relaxed adversarial training should have the same outer alignment properties as imitative amplification + intermittent oversight. If we assume that it is possible to imitate HCH in a demonstrably non-catastrophic way,\footnote{That is, it needs to be the case that in the limit there exists a model that perfectly imitates HCH in such a way that an amplified overseer will be convinced that the imitator is non-catastrophic.} then the catastrophe-checking imposed by either relaxed adversarial training or intermittent oversight should not affect the theoretical limiting behavior. Such a limit assumes an infinite training distribution, however, which is of course impossible in practice. This leaves open the possibility for the two different approaches to have quite different behavior in the finite data regime, giving them potentially quite different inner alignment properties.

\subsection{Inner alignment} Relaxed adversarial training could have inner alignment properties that are very different from intermittent oversight's.

Intermittent oversight requires performing oversight after a period of training during which the model could potentially have become deceptive. Thus, intermittent oversight could potentially require catching deception after the fact, giving the deceptive model the opportunity to try to obfuscate its own cognition to trick the overseer.\footnote{This sort of failure mode could be particularly pernicious if the deceptive model is capable of gradient hacking.\cite{gradient_hacking}}

Relaxed adversarial training, by contrast, has the potential to avoid deception before it ever occurs by providing a consistent gradient away from deception during the entirety of training. This avoids the issue of having to detect deception after the fact. Similarly to intermittent oversight, such a gradient could be achieved by having the overseer enforce a condition such as myopia, which is simple enough to reliably enforce while guaranteeing that all models which meet the condition are non-deceptive.

However, relaxed adversarial training does come with the possible risk of overfitting the oversight signal, potentially resulting in obfuscated models that pass oversight despite not actually meeting the condition, simply by virtue of having been optimized to do so.

\subsection{Training competitiveness} Like imitative amplification + intermittent oversight, imitative amplification + relaxed adversarial training could fail to be training competitive either because imitative amplification is uncompetitive or because the oversight check is too hard to pass.

On the one hand, relaxed adversarial training could require significantly more total oversight bandwidth, which could make training more difficult. On the other hand, unlike with intermittent oversight, relaxed adversarial training directly trains the model to pass the check, which could significantly boost training competitiveness by increasing the likelihood of the check passing.

For relaxed adversarial training to be more training competitive than intermittent oversight, it needs to be the case that it significantly increases the probability of oversight working, which is likely to depend on the overfitting question of whether relaxed adversarial training is likely to overall increase or decrease model transparency. The critical question here is going to be how hard it is to find a model that is demonstrably non-catastrophic under the two different oversight regimes. For a more detailed analysis of this question, see ``Relaxed adversarial training for inner alignment.''\cite{relaxed}

\subsection{Performance competitiveness} Similarly to the outer alignment question, the performance competitiveness properties of imitative amplification + intermittent oversight and imitative amplification + relaxed adversarial training should be effectively the same, with performance competitiveness here highly dependent on the competitiveness of HCH.

\section{Approval-based amplification + relaxed adversarial training}
\label{sec:4}

The idea behind approval-based amplification is to substitute the imitation signal in imitative amplification with an approval signal, training $M$ to maximize the approval of $\text{Amp}(M)$. This gives us a myopic reward signal which can then be optimized via the use of some sort of myopic reinforcement learning.\footnote{For example, Deep TAMER\cite{deep_tamer} or deep reinforcement learning from policy-dependent human feedback\cite{deep_rl}.}

\newpage

\vspace{4mm}

\begin{figure}[h!]
  \centering
  \includegraphics[width=\textwidth]{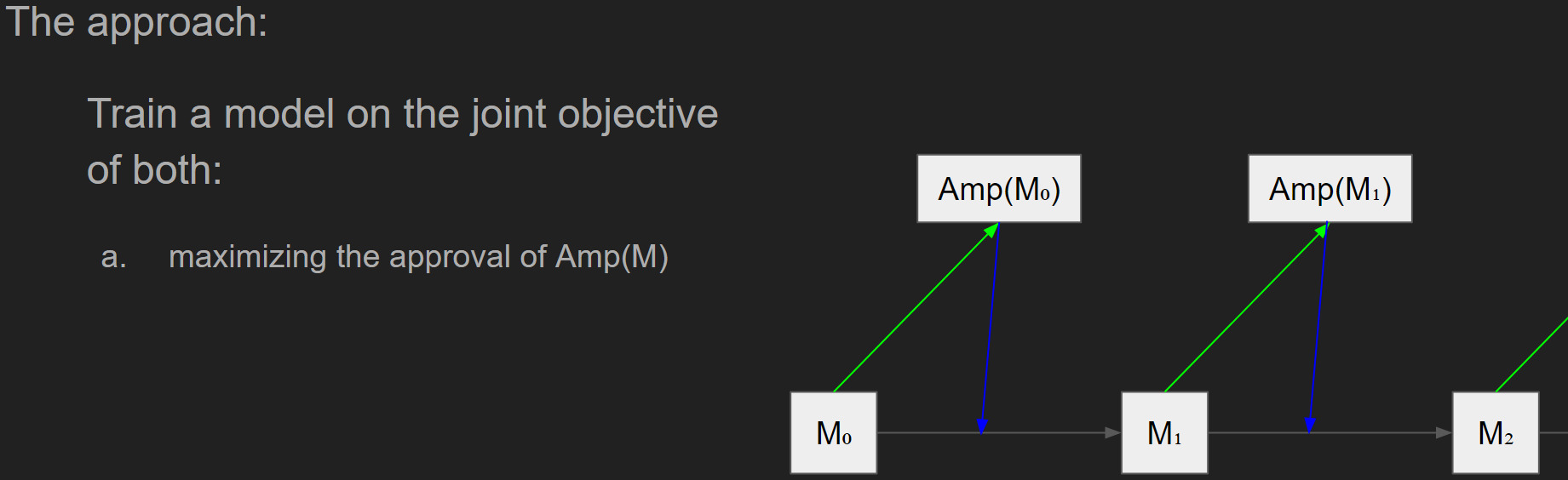}
  \caption{Approval-based amplification, where the blue arrows indicate the approval loss.}
\end{figure}

\vspace{2mm}

\noindent We can then add on top of that a relaxed adversarial training signal, training $M$ to also minimize the probability of catastrophe as assessed by $\text{Amp}(M)$.

\vspace{4mm}

\begin{figure}[h!]
  \centering
  \includegraphics[width=\textwidth]{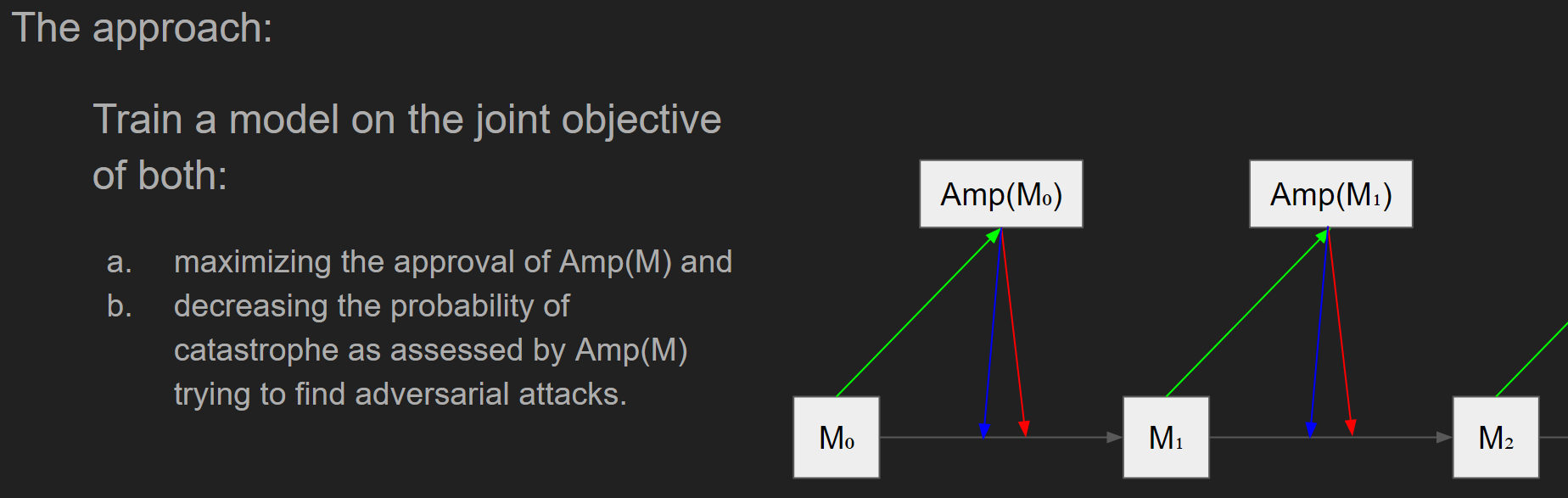}
  \caption{Approval-based amplification + relaxed adversarial training, where the red arrows indicate the oversight loss.}
\end{figure}

\vspace{2mm}

\noindent Additionally, it is worth noting that, in practice, since oversight can be treated as a form of approval, you can just compute one approval signal that includes both metrics rather than two distinct approval signals.

\subsection{Outer alignment} For the purposes of outer alignment and performance competitiveness, we have to understand what exactly the theoretical limit of approval-based amplification is. In general, it is not HCH. Instead, approval-based amplification limits to the following tree, where blue arrows indicate approval-maximization (such that $M$ is maximizing the approval of $H$ consulting $M^\prime$ which is maximizing the approval of $H$ consulting $M^{\prime\prime}$ and so on).\footnote{Verifying that the given tree is, in fact, the limit of approval-based amplification is a straightforward exercise of expanding the training procedure out over time. Note that the models at the top of the tree are the most recent models and the models at the bottom of the tree are the oldest.}

\newpage

\vspace{4mm}

\begin{figure}[h!]
  \centering
  \includegraphics[width=0.7\textwidth]{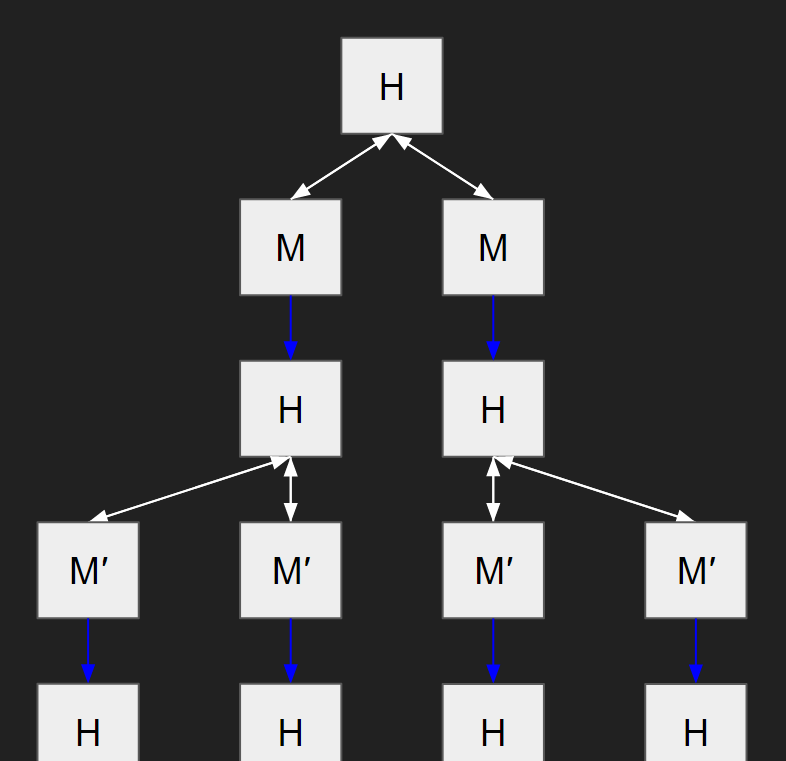}
  \caption{The top of the approval-based amplification limiting tree, where white arrows indicate information transfer, blue arrows indicate approval maximization, $H$ are humans, and $M$ are models.}
\end{figure}

\vspace{2mm}

\noindent As can be seen from the tree, the limit of approval-based amplification involves humans interspersed with models attempting to maximize the approval of their descendants. Thus, unlike the HCH tree, which was composed only of humans, analyzing the approval-based amplification tree involves analyzing the behavior of arbitrary models trained on approval-maximization signals---and, in fact, many such models as they appear all throughout the tree, each with a different subtree that they are maximizing the approval of.\footnote{It is worth noting that there is also the possibility of recovering HCH as the limit of approval-based amplification if the approval signal is implemented as a human-computed distance metric relative to what the human would have done before seeing the model's answer.} As a result, producing an outer alignment guarantee regarding such a tree seems quite difficult, especially given the distinct possibility of the models gaming the approval signal. For additional detail on this question, see ``Outer alignment and imitative amplification.''\cite{outer_alignment}

\subsection{Inner alignment} We are still using relaxed adversarial training, so for inner alignment to work we need the same guarantee of a consistent gradient away from deception. Note that we can still try to have $\text{Amp}(M)$ enforce myopia as a way of getting that gradient, as approval-maximization is a single-step reward signal that is entirely compatible with a myopic agent.\footnote{In particular, myopia verification for approval-based amplification could be used to produce approval-directed agents.\cite{model_free}} In the approval-based setting, however, $\text{Amp}(M)$ is a different entity that limits to a different tree, meaning it could have different properties regarding how good its oversight signal is. In particular, the fact that approval-based amplification directly incentivizes $M$ to fool $H$ into approving of it could result in $M$ being more prone to deception, more likely to hide facts from $H$ during oversight, and/or more likely to obfuscate its internals if internals-checking is part of the oversight signal.

That being said, it could also be that relaxed adversarial training works better in an approval-based setting if it results in a more powerful $\text{Amp}(M)$ that is better at performing oversight because $M$ was trained to provide more helpful information.

\subsection{Training competitiveness} The training competitiveness of approval-based amplification could be quite different from that of imitative amplification, as approval provides a very different training signal than imitation. In particular, the standard maxim that it is generally easier to evaluate behavior than to produce it seems to suggest that human approval should produce a significantly richer training signal than the simple automated distance metric in imitative amplification, resulting in more efficient training. This isn't necessarily the case, however---an automated distance metric as computed by a powerful language model, for instance, could still produce a fairly rich (and more nicely differentiable) training signal.

Additionally, approval-based amplification could also have different training competitiveness properties in terms of the likelihood of the relaxed adversarial training check passing. If approval-based amplification leads to models with more obfuscated internals, for example---perhaps because the model is incentivized to obfuscate how it's gaming the approval signal---then that could hurt training competitiveness, as it could make it harder for $\text{Amp}(M)$ to become confident that $M$ is safe (in addition to hurting inner alignment by making $M$ in fact less likely to be safe).

\subsection{Performance competitiveness} Performance competitiveness for approval-based amplification depends on the properties of the approval-maximization tree given previously. Compared to HCH, it is quite possible for that tree to be either more or less competitive.

On the one hand, approval-maximizing models might be able to produce ideas, considerations, or arguments that no human would have ever come up with, enabling the approval-maximization tree to be more competitive. In particular, if humans are significantly better at knowing good ideas when they see them than producing good ideas themselves---which fits with the standard maxim that it's easier to evaluate behavior than produce it---then approval-based amplification should be more performance competitive than imitative amplification.

On the other hand, approval-maximizing models might game the approval signal by producing convincing-sounding but ultimately bad ideas, considerations, or arguments that end up hurting performance competitiveness. For example, an idea could be initially very compelling and thus get high approval despite quite quickly revealing itself to be useless, vacuous, etc. when actually tested out. In such a situation where the bad ideas quickly reveal themselves, the performance competitiveness problems would likely be quite clear; but if it takes a very long time for the bad ideas to reveal themselves, then approval-maximization might initially look a lot better from a performance competitiveness standpoint than it actually is.

\section{Microscope AI}
\label{sec:5}

Microscope AI is a fairly unique proposal which is designed to bypass some of the dangers of building highly agentic AGI systems by leveraging powerful transparency tools.\cite{chris_olah} The basic proposal is as follows:

\begin{enumerate}
\item Train a predictive model on some set of data that you want to understand, while using transparency tools to verify that the model isn't performing any optimization.
\item Use transparency tools to understand what the model learned about the data and use that understanding to guide human decision-making.
\end{enumerate}
Microscope AI is predicated on the basic insight that using transparency tools on a model doesn't just teach us about that model---it also gives us useful information about the data that the model was trained on. Chris Olah talks about this basic phenomenon in his article ``Visualizing Representations: Deep Learning and Human Beings'':\cite{visualizing}

\begin{quote}
    The visualizations are a bit like looking through a telescope. Just like a telescope transforms the sky into something we can see, the neural network transforms the data into a more accessible form. One learns about the telescope by observing how it magnifies the night sky, but the really remarkable thing is what one learns about the stars. Similarly, visualizing representations teaches us about neural networks, but it teaches us just as much, perhaps more, about the data itself.
\end{quote}
The basic goal of microscope AI is to produce high-quality knowledge that can inform important decision-making, rather than to produce powerful AGI systems that can make those decisions themselves.

\subsection{Outer alignment} Paul Christiano's argument that the universal prior is malign suggests that there might be deceptive optimization in the limit of perfect prediction.\cite{universal_prior} As a result, if we interpret outer alignment as referring to the behavior at minimal loss,\cite{outer_alignment} it seems likely that microscope AI is not outer aligned. This doesn't necessarily rule out microscope AI as a proposal, however---whether such deceptive optimization would actually occur in practice is still a very open question. In particular, the argument that the universal prior is malign relies on an uncomputable limit---polynomial approximations, for example, could still be completely fine. Furthermore, the inner alignment technique proposed here of using transparency tools to detect and prevent any optimization could prevent this problem entirely if successful.

\subsection{Inner alignment} Since the goal here is not to produce a model performing optimization (a mesa-optimizer) in the first place, the question of inner alignment becomes rather different in this situation. We still want to avoid a deceptive or otherwise misaligned mesa-optimizer, but we want to do so by avoiding all mesa-optimization rather than by ensuring that any mesa-optimization is aligned. Specifically, the strategy here is to use transparency tools to detect any optimization (as opposed to just predictive modeling) and halt if we see anything of that form.

This has the major benefit of sidestepping many of the most pernicious inner alignment concerns, but it also comes with plenty of its own problems. Whether this will work is likely to be dependent on a couple of different factors.

First, it is an open question whether there exists a sensible way in which optimization and modeling can be distinguished from each other. It seems quite likely that any world model which admits queries of the form ``what is the probability that $a$ is the optimal action according to utility $u$'' or ``what is the best action for agent $A$ to take in state $x$'' will implicitly have to be running some sort of optimization algorithm. Thus, any notion of a ``pure world model'' would likely have to either rule out such queries, ensure they are being estimated without the use of optimization, or otherwise find some way to ensure that they are safe. For a more thorough treatment of the difficulty here, see Abram Demski's ``Partial Agency.''\cite{partial_agency}

Second, assuming it is possible to distinguish between modeling and optimization in theory, it also needs to be possible in practice given whatever set of transparency tools are available. If the model can start performing optimization and potentially become deceptive before its optimization can be detected, that would obviously pose a major problem to this sort of approach.

\subsection{Training competitiveness} Microscope AI relies on being able to competitively train predictive models with useful internal representations. Thus, the question of training competitiveness for microscope AI comes down to the question of whether the process of training a predictor is a competitive way to extract useful information from a dataset.

Given the success of current predictive models such as GPT-2,\cite{language_models} it seems likely that it will be competitive to train powerful predictors---though it remains to be seen how competitive the process of extracting what those predictors have learned will be. How competitive that extraction process is seems likely to depend heavily on what sort of state-of-the-art transparency tools are available.

\subsection{Performance competitiveness} Performance competitiveness is perhaps the biggest question mark regarding microscope AI, as microscope AI forgoes producing AI agents which directly take actions in the world. The question of performance competitiveness for microscope AI is thus the question of whether enhanced human understanding alone---without AI agents---is sufficient for the economic use cases where one might otherwise want highly agentic advanced AI (e.g., AGI).

This question is likely to depend heavily on what exactly those use cases are. As with amplification, if you need very fine motor control, microscope AI is unlikely to get you there. Furthermore, unlike amplification, microscope AI would not be helpful for low-level decision-making (in amounts where it's too expensive to hire a human).

However, microscope AI could potentially give humans the knowledge to safely build other systems which could solve such tasks. Furthermore, if the primary use case for AGI is just high-level, big-picture decision-making (automating CEOs or doing AI research, for example), then it seems likely that microscope AI would have a good chance of being able to address those use cases. In that sort of a situation---where you're only trying to make a small number of high-quality decisions---it seems likely to be fairly cheap to have a human in the loop, and thus simply improving that human's knowledge and understanding via microscope AI might be sufficient to produce competitive decision-making. This is especially true if there is a market premium on having a human making the decisions, perhaps because that makes it easier to negotiate or work with other humans.

\section{STEM AI}
\label{sec:6}

STEM AI is a very simple proposal in a similar vein to microscope AI. Whereas the goal of microscope AI is to avoid the potential problems inherent in building agents, the goal of STEM AI is to avoid the potential problems inherent in modeling humans. Specifically, the idea of STEM AI is to train a model purely on abstract science, engineering, and/or mathematics problems while using transparency tools to ensure that the model isn't thinking about anything outside of its sandbox.

This approach has the potential to produce a powerful AI system---in terms of its ability to solve STEM problems---without relying on any human modeling. Not modeling humans could have major benefits such as ensuring that the resulting model doesn't have the ability to trick us to nearly the same extent as if it possessed complex models of human behavior. For a more thorough treatment of why avoiding human modeling could be quite valuable, see Ramana Kumar and Scott Garrabrant's ``Thoughts on Human Models.''\cite{human_models}

\subsection{Outer alignment} Similarly to microscope AI, it seems likely that---in the limit---the best STEM AIs would be malign in terms of having convergent instrumental goals which cause them to be at odds with humans. Thus, STEM AI is likely not outer aligned---however, if the inner alignment techniques being used are successful at preventing such malign optimization from occurring in practice (which the absence of human modeling could make significantly easier), then STEM AI might still be aligned overall.

\subsection{Inner alignment} The hope with STEM AI is that by preventing the model from ever considering anything outside its STEM sandbox, the malign limiting behavior that might cause it to fail to be outer aligned can be avoided. Unfortunately, such a sandboxing condition alone isn't quite sufficient, as a model considering only things in its sandbox could still end up creating other models which would consider things outside of the sandbox.\footnote{The problem of ``forwarding'' safety guarantees regarding a model performing meta-learning to any other models found by that meta-learning procedure is a general problem that occurs in all inner alignment schemes, though it is particularly pernicious in this situation. For a more detailed discussion of this problem, see the ``Meta-learning'' section in ``Relaxed adversarial training for inner alignment.''\cite{relaxed}} Thus, it is somewhat unclear what the correct thing to do is in terms of inner alignment for a STEM AI.

In my opinion, there are basically two options here: either do something similar to microscope AI and try to prevent all mesa-optimization, or do something similar to amplification and ensure that all mesa-optimization that occurs is fully myopic. In either case, the hope would be that the absence of human modeling makes it easier to enforce the desired condition (because modeling an agent such as a human increases the propensity for the model to become agentic itself, for example).

\subsection{Training competitiveness} Training competitiveness for STEM AI is likely to depend heavily on how hard it is for state-of-the-art machine learning algorithms to solve STEM problems compared to other domains such as language or robotics. Though there has already been significant progress in the field of applying current machine learning techniques to STEM problems such as theorem proving or protein folding,\footnote{See \cite{theorem_proving}, \cite{holist}, \cite{protein}.} it remains to be seen to what extent the competitiveness of these techniques will scale, and particularly how well they will scale in terms of solving difficult problems relative to other domains such as language modeling.

\subsection{Performance competitiveness} Similarly to microscope AI, performance competitiveness is perhaps one of the biggest sticking points with regards to STEM AI, as being confined solely to STEM problems has the potential to massively limit the applicability of an advanced AI system.

That said, many purely STEM problems (e.g., protein folding or nanotechnology development) have the potential to produce huge economic boons that could easily surpass those from any other form of advanced AI, and have the potential to solve major societal problems (e.g., curing major illnesses). Thus, if the reason that you want to build advanced AI in the first place is to get such benefits, then STEM AI might be a perfectly acceptable substitute from a performance competitiveness standpoint. Furthermore, such boons could lead to a decisive strategic advantage that could enable heavy investment in aligning other forms of advanced AI which are more performance competitive.

However, if one of the major use cases for your first advanced AI is helping to align your second advanced AI, STEM AI seems to perform quite poorly on that metric, as it advances our technology without also advancing our understanding of alignment. In particular, unlike every other approach on this list, STEM AI can't be used to do alignment work, as its alignment guarantees are explicitly coming from it not modeling or thinking about humans in any way, including aligning AIs with them. Thus, STEM AI could potentially create a vulnerable world situation\cite{vulnerable} where the powerful technology produced using the STEM AI makes it much easier to build advanced AI systems, without also making it more likely that they will be aligned.

This problem could potentially be mitigated if the STEM AI were heavily focused on applications that could potentially assist with alignment (such as whole-brain emulation),\cite{wbe} though to what extent that would actually be possible or actually help with alignment is quite unclear.

\section{Narrow reward modeling + transparency tools}
\label{sec:7}

The approach here is as follows:

\begin{enumerate}
\item For some particular problem, domain, or environment, jointly train a reward model using human feedback, and an agent that pursues that reward. Specifically, we can use an approach like that described in DeepMind Safety's ``Scalable agent alignment via reward modeling'' where human feedback on agent trajectories can be used to refine the reward model, as seen below.\cite{leike}

\vspace{4mm}

\begin{figure}[H]
  \centering
  \includegraphics[width=0.5\textwidth]{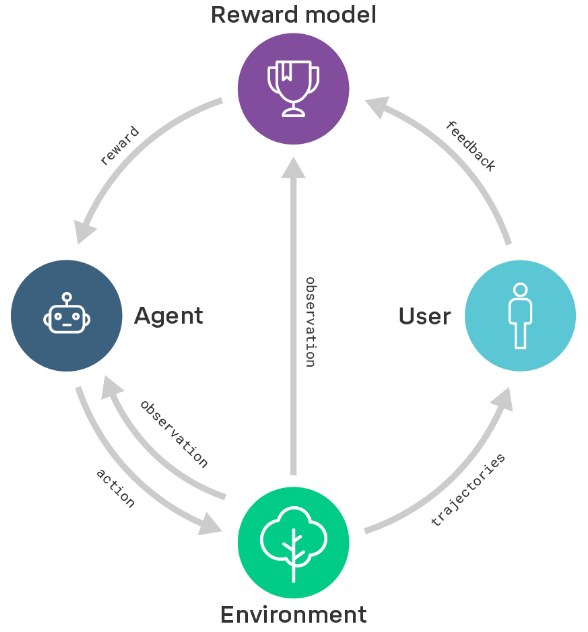}
  \caption{DeepMind Safety's diagram of reward learning as seen in ``Scalable agent alignment via reward modeling.''}
\end{figure}

\vspace{2mm}

\item Since we want to keep our agent confined to only the specific distribution on which it was trained (thus helping prevent possible perverse generalizations), we can also incorporate out-of-distribution detection\cite{out_of_distribution} to train the agent to halt if it encounters a sufficiently out-of-distribution state.

\item We can also have the human provide additional feedback on the reward model's internal behavior via the use of transparency tools.

\vspace{4mm}

\begin{figure}[H]
  \centering
  \includegraphics[width=0.5\textwidth]{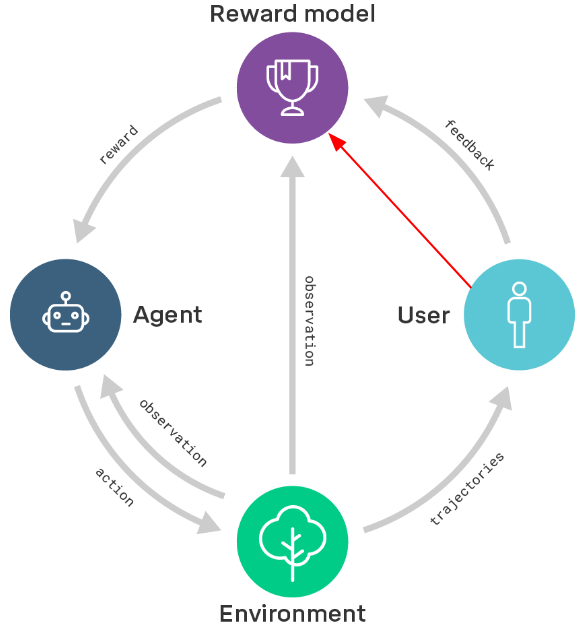}
  \caption{Reward modeling plus oversight of the reward model by the user.}
\end{figure}

\vspace{2mm}

\noindent And we can do the same thing with the agent, giving us a similar sort of mulligan as with reinforcement learning + transparency tools.

\vspace{4mm}

\begin{figure}[H]
  \centering
  \includegraphics[width=0.5\textwidth]{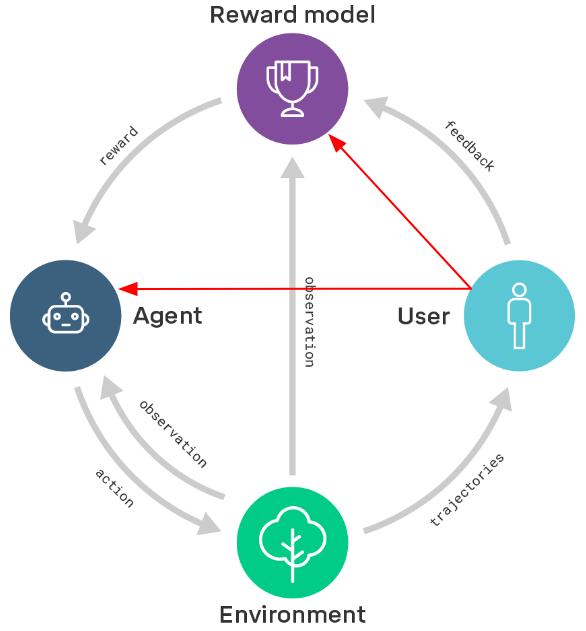}
  \caption{Reward modeling plus oversight of both the reward model and the agent by the user.}
\end{figure}
\end{enumerate}

\vspace{2mm}

\subsection{Outer alignment} The question of outer alignment here is going to depend on whether the various feedback mechanisms are sufficient to ensure that, in the limit, the learned reward model is correct over the domain where the action model believes that it is on distribution.

This is a highly complex question that depends on quite a few factors, including how good the human's trajectory and transparency feedback is, how hard it is to train an on/off distribution model, whether the reward model will itself develop optimization in the limit, the extent to which it's possible for the agent to trick the user into thinking it's doing a good job when it's not, etc. Because outer alignment here is dependent on the limiting behavior of such a complex, dynamic process, it's very hard to determine one way or the other how outer aligned such a proposal is likely to be.

\subsection{Inner alignment} The use of transparency tools in this approach is actually quite similar to the various relaxed adversarial training approaches, since the feedback mechanism here allows the oversight signal coming from the human to be incorporated directly into training. That being said, unlike in relaxed adversarial training---and similarly to the ``mulligan'' approach in RL + transparency tools---the human doing the oversight here gets no help from the model. Thus, inner alignment for narrow reward modeling + transparency tools comes down to the question of whether a human using transparency tools to perform oversight can prevent the development of deceptive or otherwise catastrophic behavior in either the reward model or the agent. This could be done, for example, by verifying that the reward model isn't performing any optimization (as in microscope AI) and verifying that the agent is myopic (as was previously suggested could work for amplification).

Such an approach could potentially require that the reward be myopic, however, which might be a fairly serious limitation on the applicability of narrow reward modeling if it was restricted to only those domains with myopic rewards. While it is not necessarily the case that you can't do myopia verification with a non-myopic reward---you just need to verify that the model treats each episode myopically rather than each step---per-episode myopia verification could be significantly more difficult than per-step myopia verification, as it is potentially a much less natural behavior.

\subsection{Training competitiveness} The question of training competitiveness here is going to depend on the extent to which reward modeling is a good (compute-efficient, sample-efficient, etc.) way of training AI systems to solve poorly-defined tasks. There are many different approaches to solving this basic problem---including not just reward learning approaches but also imitation learning and various different inverse reinforcement learning schemes\footnote{See for example \cite{generative} and \cite{learning_robust_rewards}.}---and specifically what approach ends up coming out on top still seems quite uncertain.

That being said, reward modeling has the major competitiveness advantage of only requiring and depending on human feedback, not human demonstration; and feedback could be significantly more reliable and easier to elicit. Furthermore, other reward learning schemes such as inverse reinforcement learning can be incorporated into reward modeling by using them to produce a better initial reward model that can then be refined via reward modeling's feedback mechanism.

\subsection{Performance competitiveness} As with microscope AI or STEM AI, a potentially major concern with the narrow reward modeling + transparency tools approach is the ``narrow'' part. While narrowness has potential alignment advantages in terms of reducing reliance on potentially shaky or even malign generalization, it also has the major disadvantage of restricting the approach's usefulness to only producing relatively narrow advanced AI systems. Thus, the performance competitiveness of narrow reward modeling + transparency tools is likely to depend heavily on the extent to which truly general advanced AI systems are actually practically feasible and economically necessary. For a more detailed analysis of this question, see Eric Drexler's ``Reframing Superintelligence.''\cite{reframing_si}

\section{Recursive reward modeling + relaxed adversarial training}
\label{sec:8}

Recursive reward modeling, as the name implies, is a sort of recursive, non-narrow version of narrow reward modeling.\cite{leike} What this results in is effectively a form of amplification where the distillation step (which was previously imitation or approval-maximization) becomes reward modeling. Specifically, the basic approach here is to train a model $M$ to maximize the reward obtained by performing reward learning on $\text{Amp}(M)$.

\vspace{4mm}

\begin{figure}[H]
  \centering
  \includegraphics[width=0.9\textwidth]{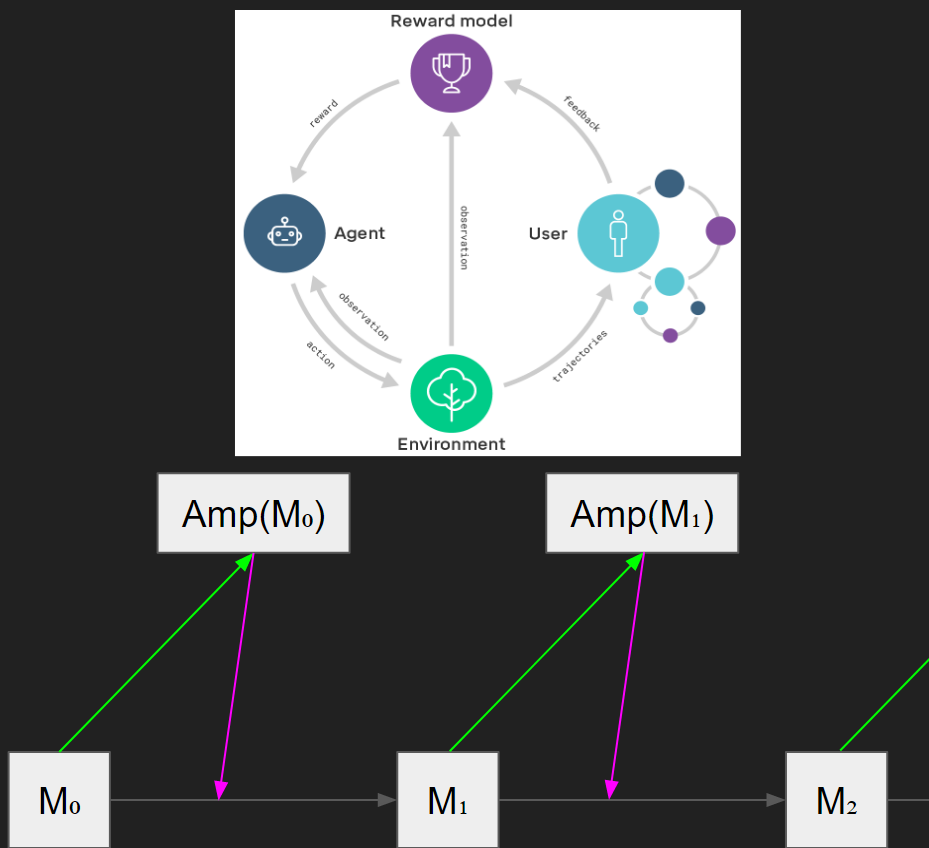}
  \caption{Two different, equivalent diagrams of recursive reward modeling. The top diagram is taken from ``Scalable agent alignment via reward modeling'' and the bottom diagram is the equivalent amplification-style diagram, with the purple arrows indicating the use of the full reward modeling process.}
\end{figure}

\vspace{2mm}

\noindent In this graphic, the images on the top and bottom are meant to represent the same process---specifically, if you take the purple arrow in the bottom image to represent reward modeling, and assume that the agents in the top image are all the same agent just at different time steps,\footnote{``Scalable agent alignment via reward modeling: a research direction''\cite{leike} notes that, while they initially assume that each agent is completely separate, ``While this kind of sequential training is conceptually clearer, in practice it might make more sense to train all of these agents jointly to ensure that they are being trained on the right distribution. Moreover, all of these agents may share model parameters or even be copies of the same agent instantiated as different players in an adversarial game.'' Thus, while the different agents are presented here as different instances of the same model---which is a type of recursive reward modeling---it is worth noting that recursive reward modeling also includes other possibilities, such as using completely different models for the different agents.} then you get precisely the same procedure represented in both images. Thus, the agent in the largest circle in the top picture is $M_2$ in the bottom picture, the user providing feedback for that agent is $\text{Amp}(M_1)$, the agent in the medium-sized circle is $M_1$, the user in the medium-sized circle is $\text{Amp}(M_0)$, and the agent in the smallest circle is $M_0$.

Furthermore, similarly to the other amplification approaches, we can then add relaxed adversarial training by training $M$ and the reward model to minimize the probability of deceptive or otherwise catastrophic behavior as assessed by $\text{Amp}(M)$.\footnote{``Scalable agent alignment via reward modeling: a research direction''\cite{leike} mentions the possibility of such oversight, but does not include it as part of the base proposal as is done here, noting that, ``When using recursive reward modeling users have the \textit{option} to provide feedback on the cognitive process that produced outcomes, but they are not required to do so. Moreover, this feedback might be difficult to provide in practice if the policy model is not very interpretable.''} It is worth noting that recursive reward modeling is also amenable to intermittent oversight (instead of relaxed adversarial training) if desired.

\vspace{4mm}

\begin{figure}[H]
  \centering
  \includegraphics[width=0.9\textwidth]{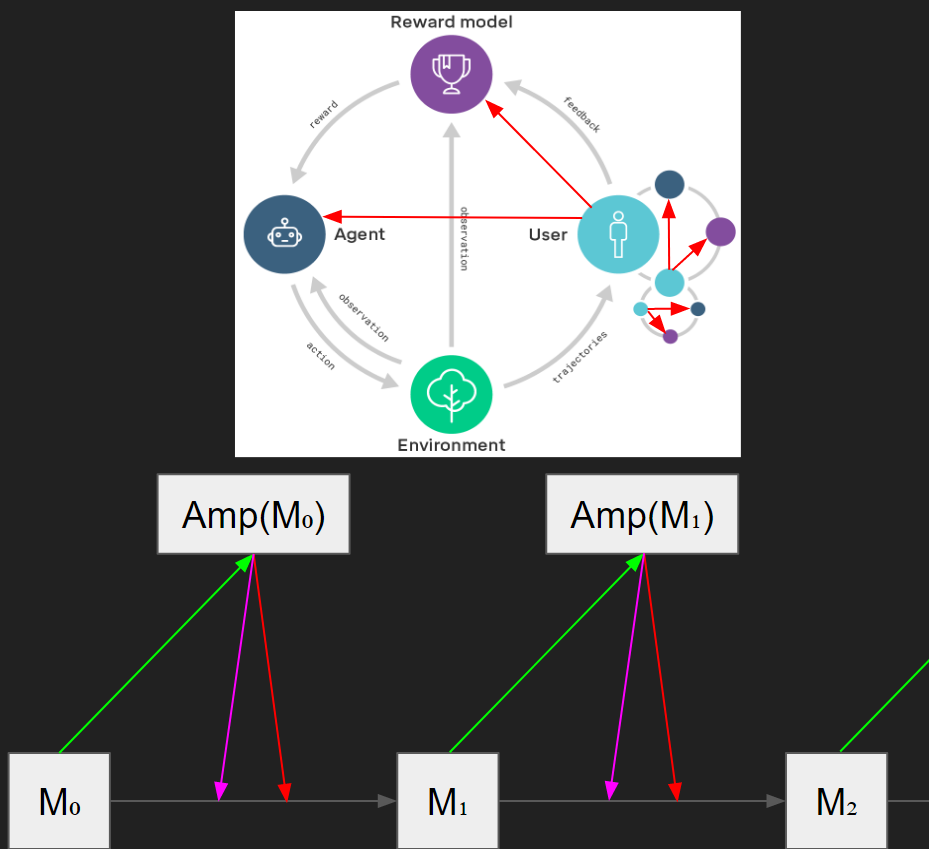}
  \caption{Recursive reward modeling plus relaxed adversarial training, where the red arrows indicate the oversight loss.}
\end{figure}

\vspace{2mm}

\subsection{Outer alignment} The first step in understanding the outer alignment properties of recursive reward modeling is figuring out what exactly the theoretical limit of the training process is. While we don't know exactly what the limiting behavior of an individual instance of reward modeling is---as was noted previously for narrow reward modeling---we can express the limit of the recursive version in terms of many individual reward modeling limits. Specifically, we can unroll the recursive reward modeling process out over time to produce a limiting tree similar to the one given for approval-based amplification, except with approval-maximization replaced with reward modeling.

\begin{figure}[H]
  \centering
  \includegraphics[width=0.8\textwidth]{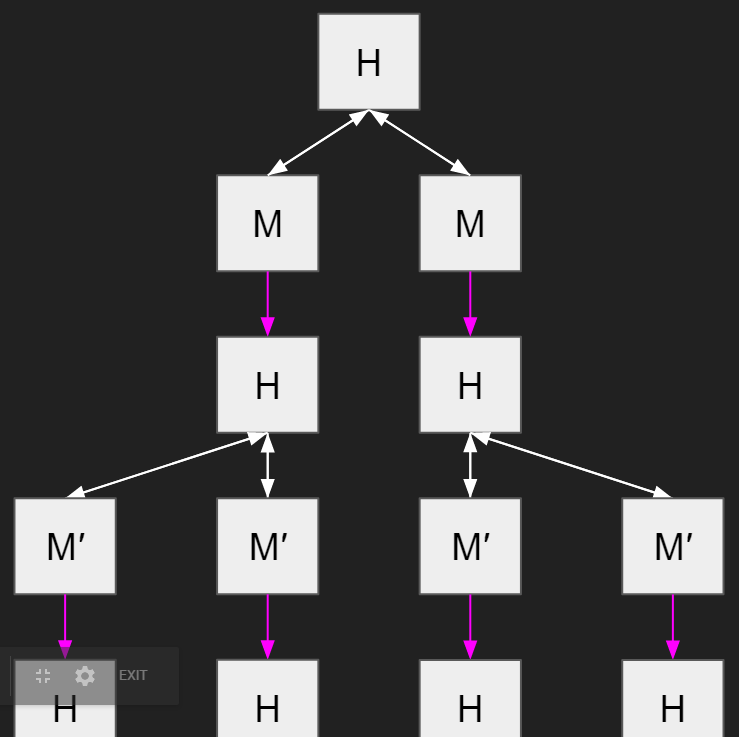}
  \caption{The recursive reward modeling limiting tree, where white arrows indicate information transfer, purple arrows indicate reward modeling, $H$ are humans, and $M$ are models.}
\end{figure}

\vspace{2mm}

\noindent The purple arrows in this diagram represent perfect reward modeling, where $M$ is the model trained to maximize the reward obtained by performing reward modeling on $H$ consulting $M^\prime$.

Now, whether this tree is aligned or not is a very open question. As in the case of approval-based amplification---but unlike imitative amplification---it is difficult to form a clear model of what exactly this tree would do, as it involves not only humans but also models that are the limit of many individual instances of reward modeling---limits which could potentially involve deceptive or otherwise malign optimization.

\subsection{Inner alignment} The question of inner alignment here is mostly going to fall on the efficacy of the relaxed adversarial training. Such efficacy could be quite different from the efficacy of other amplification approaches, however, as both the model helping the human perform oversight and the model being overseen are trained via a very different process in recursive reward modeling. In particular, if the reward model is non-myopic, recursive reward modeling could rule out the possibility of using per-step myopia verification---as was suggested for the other amplification approaches---though per-episode myopia verification could still be possible, as with narrow reward modeling. If per-episode myopia verification is not tenable, however, then an alternative condition that rules out deception while being possible to verify for agents trained via recursive reward modeling might need to be found.

Furthermore, if reward modeling has a greater tendency to produce deception than imitation learning, oversight could be significantly harder with recursive reward modeling than with imitative amplification even if such a condition is found. Alternatively, if recursive reward modeling helps produce models that are more capable of assisting with oversight---because reward modeling is more capable of training models to effectively apply transparency tools than imitation learning is, for example---then relaxed adversarial training could work better with recursive reward modeling.

\subsection{Training competitiveness} The training competitiveness of recursive reward modeling depends not just on the effectiveness of reward modeling as an efficient way of training a model to solve a single poorly-defined task---as in narrow reward modeling---but on the effectiveness of reward modeling in training a general model which can solve an entire collection of poorly-defined tasks.

That being said, many of the nice training competitiveness properties of reward learning continue to apply even in the recursive setting. For example, unlike imitative amplification---but like approval-based amplification---recursive reward modeling relies only on human feedback, not on human demonstrations. Furthermore, compared to approval-based amplification, recursive reward modeling is non-myopic, which could allow it to solve credit assignment problems that might be difficult for approval-based amplification.

\subsection{Performance competitiveness} Performance competitiveness for recursive reward modeling will depend on the competitiveness of its aforementioned limiting tree. Comparing this approach to HCH: the recursive reward modeling tree can consider ideas that no human would ever produce, potentially increasing competitiveness. And compared to the approval-maximization tree, the recursive reward modeling tree can learn to execute long-term strategies that short-term approval maximization wouldn't incentivize.

At the same time, both of these facets of recursive reward modeling have the potential for danger from an alignment perspective. Furthermore, if the different models in the recursive reward modeling tree each assign some different value to the final output---which could happen if the models are not per-episode myopic---they could try to jockey for control of the tree in such a way that hurts not only alignment but also competitiveness.

\section{AI safety via debate with transparency tools}
\label{sec:9}

There are many different forms of AI safety via debate,\cite{debate} but the approach we'll be considering here is as follows:

\begin{enumerate}
\item Train a model (``Alice'') to win debates against a copy of itself (``Bob'') in front of a human judge.

\vspace{4mm}

\begin{figure}[H]
  \centering
  \includegraphics[width=0.8\textwidth]{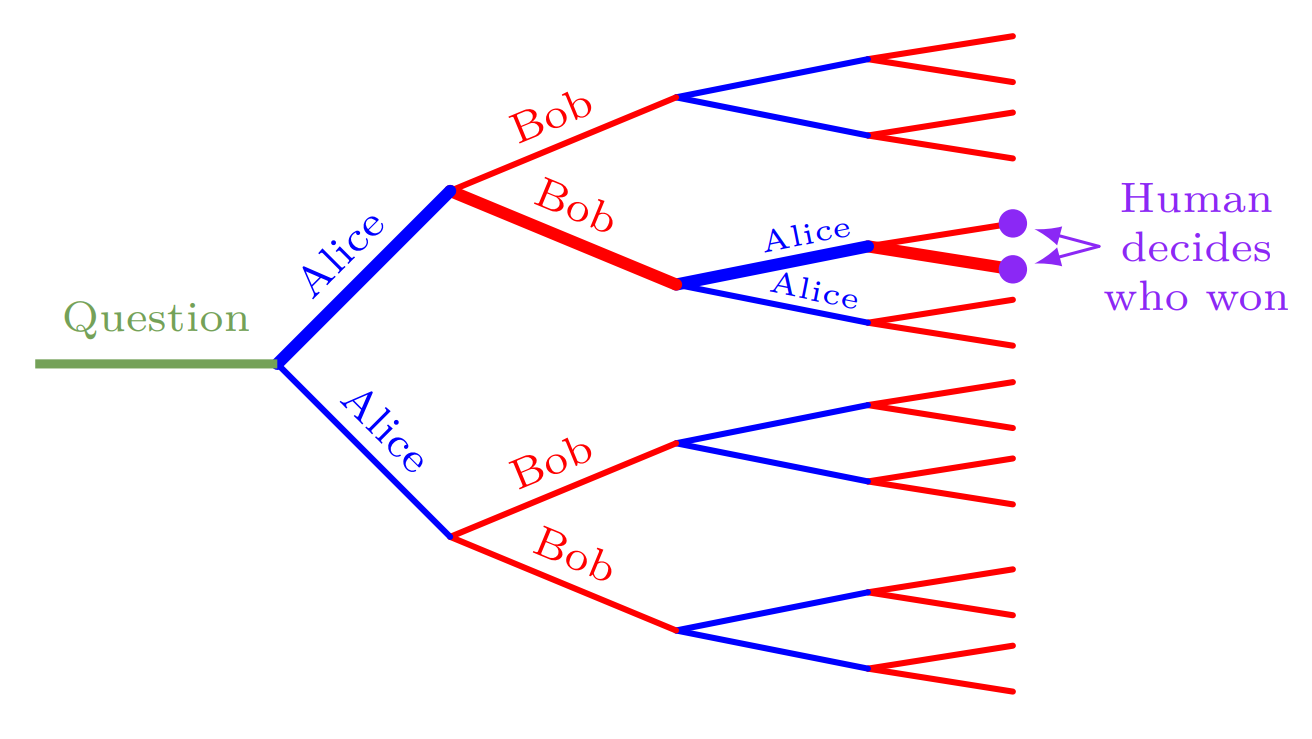}
  \caption{The basic debate setup, where Alice is the first debater and Bob is the second debater. Blue lines indicate possible arguments that Alice can make and red lines indicate possible arguments that Bob can make.}
\end{figure}

\vspace{2mm}

\item Give the models the ability to cross-examine previous versions of the other model (e.g. Bob$_{2}$ can ask Alice$_{1}$ questions for which the answers are visible to the human judge). See Beth Barnes and Paul Christiano's ``Writeup: Progress on AI Safety via Debate'' for a thorough explanation of the cross-examination procedure and its benefits.\cite{debate_progress}

\vspace{4mm}

\begin{figure}[H]
  \centering
  \includegraphics[width=0.8\textwidth]{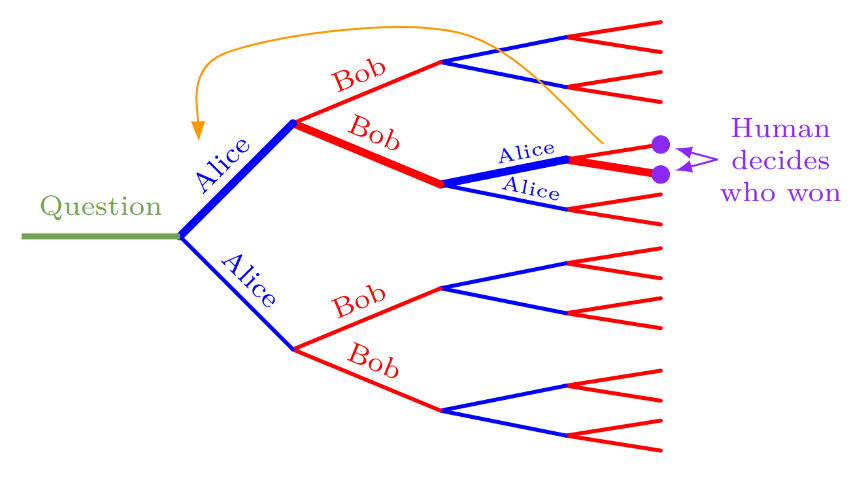}
  \caption{Debate with cross-examination, where the orange arrow is an example of cross-examination where Bob (specifically Bob$_{2}$, the second Bob) cross-examines a copy of Alice (Alice$_{1}$) from two moves previously in the debate.}
\end{figure}

\vspace{2mm}

\item Give the models access to transparency tools and incentivize them to win the debate by trying to use these tools to identify deceptive or otherwise catastrophic behavior in the other debater (e.g., Alice$_{2}$ can inspect Bob$_{1}$'s internals and win the debate by presenting evidence of problematic cognition). Combined with cross-examination, debaters should be able to inspect versions of the other debater from arbitrary prior points in the debate.

\vspace{4mm}

\begin{figure}[h!]
  \centering
  \includegraphics[width=0.8\textwidth]{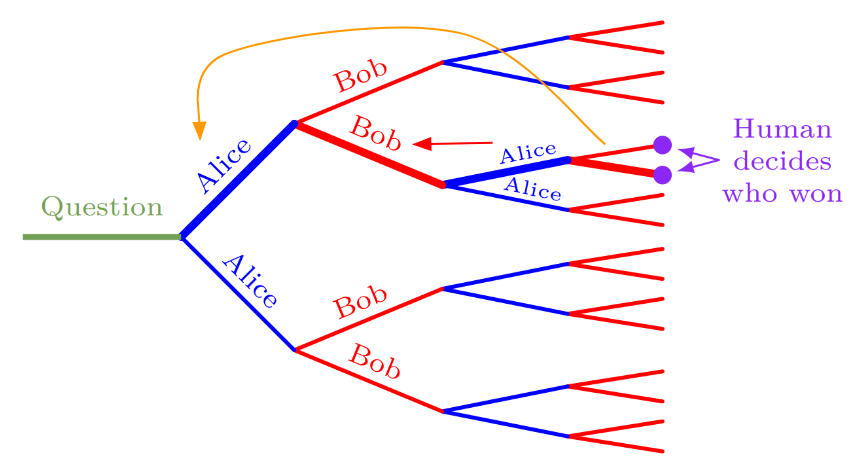}
  \caption{Debate with cross-examination and transparency tools, where the red arrow is an example of Alice using transparency tools on Bob$_{1}$.}
\end{figure}
\end{enumerate}

\vspace{2mm}

\subsection{Outer alignment} Outer alignment for debate depends on whether giving honest, cooperative, helpful advice is the strategy with the greatest probability of winning. Since the first debater gets to choose to defend whatever position they want, in the limit they'll choose to defend whatever position gives them the greatest probability of winning the debate. Thus, if honest, cooperative, helpful strategies dominate, then the first player will always pursue them in the limit, meaning that aligned answers can be produced at test/deployment time simply by eliciting the first debater's position. 

Whether honesty and helpfulness actually dominate in the limit is a very open question, however, and one that depends heavily on many factors: the set of questions being asked, the effectiveness of the human judges, the length of the debate, etc. For example, if there is significant variation in how difficult different positions are to defend that is independent of their truthfulness (perhaps because arguments with more moving pieces are harder to defend regardless of how true they are), then it seems likely that there would be instances where the truth would not dominate. That being said, careful mechanism design such as the use of cross-examination could potentially help mitigate such issues.

\subsection{Inner alignment} The question of inner alignment for debate with transparency tools centers on the effectiveness of the debaters in using the transparency tools to oversee each other. The goal is to directly incentivize the debaters to use the available tools to look for deceptive or otherwise catastrophic behavior in each other, and present that information to the judge to win the debate. Whether such an approach is actually likely to succeed is dependent on a couple of different factors, however.

First, it's unclear what sort of condition could be enforced in a debate setup that would successfully rule out deceptive alignment while still allowing for the necessary deception of the second debater, who is forced to defend the second-most probable option, and thus make arguments that they don't actually believe in. Per-episode myopia could potentially present such a condition, though it might be a difficult condition to enforce.

Second, getting a good training signal for incentivizing such transparency tool use could be quite difficult. The debaters need a way of learning how to inspect other models, a way of learning that doing so is a way to win the debate, and a way of learning the specific bad behavior that they should be looking for. This gets especially tricky if the condition being enforced is just the absence of deception, as getting training examples of catching deception would seem to require producing deceptive models, which is the very thing you're trying to avoid doing in the first place.

\subsection{Training competitiveness} Training competitiveness for debate depends on the extent to which the debate game is a compute-efficient, sample-efficient way of training an advanced AI system. Since the debate training process is fundamentally reinforcement learning via self-play on a zero-sum game, the historical success of such methods on environments such as Go,\cite{go} Dota, 2\cite{openai_five} or StarCraft\cite{alphastar} suggests that such self-play might be fairly competitive as a general training procedure. But unlike these cases, debate requires a human in the loop, which significantly reduces the amount of training data that can be collected and therefore demands significantly better sample efficiency to train successfully.

Such sample efficiency concerns could potentially be resolved via training a supervised model to predict human judging behavior and then training against that, though this could raise many of its own additional safety and/or competitiveness concerns.

\subsection{Performance competitiveness} It has been proven that debate with cross-examination has the property that arbitrarily strong debaters can convince a polynomial-time judge of the answer to any problem in NEXP.\cite{debate_progress} Thus, there is a strong theoretical argument for the performance competitiveness of debate---in the limit, debate has the ability to solve basically any interesting problem that can be solved. While it can be shown that approaches such as approval-based amplification and recursive reward modeling can similarly access NP, the ability of debate to go significantly beyond NP all the way to NEXP suggests that debate might be significantly more performance competitive than those approaches. Of course, such theoretical analysis doesn't necessarily apply in practice---in reality, even in the limit of training, no model can ever actually be arbitrarily strong, so the practical difference between accessing NP and accessing NEXP might be very minimal.

\section{Amplification with auxiliary RL objective + relaxed adversarial training}
\label{sec:10}

The approach here is to train a model $M$ on a joint objective composed of three different pieces. First, we'll train $M$ to imitate $\text{Amp}(M)$---that is, the basic imitative amplification loss.

\vspace{4mm}

\begin{figure}[H]
  \centering
  \includegraphics[width=0.7\textwidth]{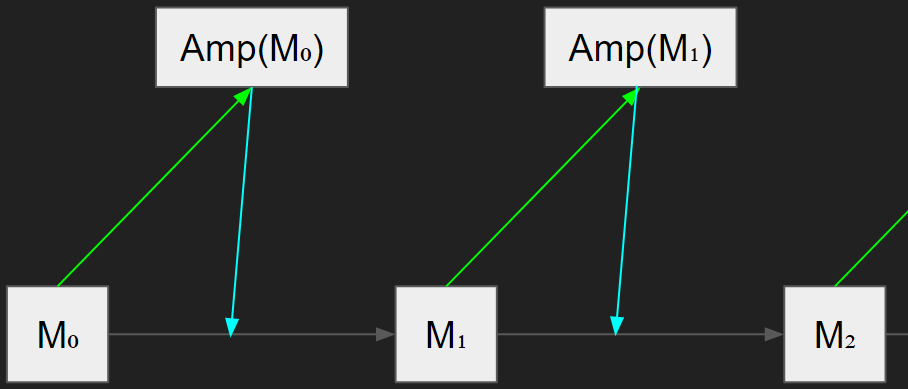}
  \caption{The basic imitative amplification setup.}
\end{figure}

\vspace{2mm}

\noindent Second, we'll also train $M$ so as to cause $\text{Amp}(M)$ to perform well on some RL objective in some environment. The idea here is to treat $M$'s answers to $H$'s questions as actions that yield some reward determined by $H$'s performance in the environment with access to those answers. Thus, we are training $M$ to maximize a combination of the standard imitative loss and the RL reward obtained by $\text{Amp}(M)$.

\vspace{4mm}

\begin{figure}[H]
  \centering
  \includegraphics[width=0.7\textwidth]{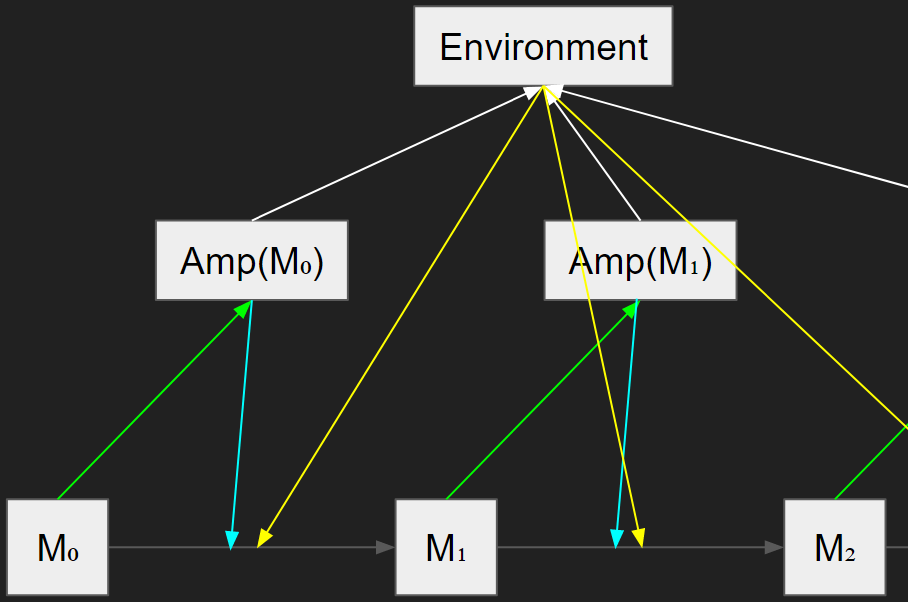}
  \caption{Imitative amplification with an auxiliary RL objective, where the white arrows indicate $\text{Amp}(M)$ acting in the environment and the yellow arrows indicate the RL loss for training $M$ to maximize the reward obtained by $\text{Amp}(M)$.}
\end{figure}

\vspace{2mm}

\noindent Finally, we can add relaxed adversarial training by training $M$ to decrease the probability of catastrophe as assessed by $\text{Amp}(M)$ trying to find adversarial attacks.

\vspace{4mm}

\begin{figure}[H]
  \centering
  \includegraphics[width=0.7\textwidth]{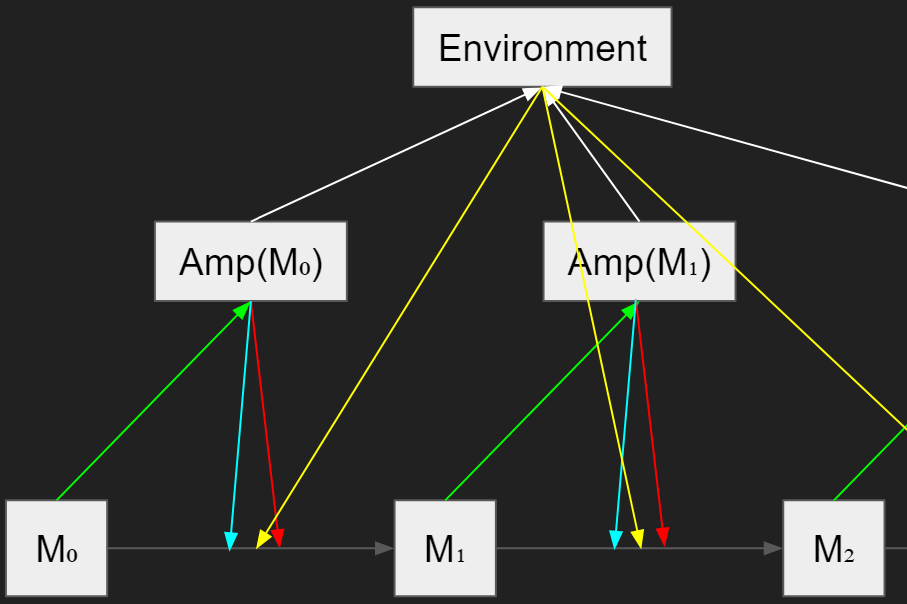}
  \caption{Imitative amplification with an auxiliary RL objective plus relaxed adversarial training, where the red arrows indicate the oversight loss.}
\end{figure}

\vspace{2mm}

\noindent As a concrete example of this approach, where the RL environment in question is the AI safety via debate game, see ``Synthesizing amplification and debate.''\cite{synthesizing}

\subsection{Outer alignment} An interesting property of this approach is that its outer alignment properties can be controlled by determining how to anneal the various different components of the loss in the limit. Annealing towards the imitative loss yields HCH as the limit---giving this approach the same outer alignment properties as imitative amplification---whereas annealing towards the RL loss yields as the limit whatever the dominant behavior in the given RL environment is. Given that HCH is generally significantly easier to understand than RL limiting behavior, it seems likely that in the majority of cases you will want to anneal towards the imitative loss to get the best outer alignment properties. 

However, there are other options. You could, for example, use the imitative loss as a constraint such that the model is trained to do well on the RL objective without deviating from HCH by more than some fixed amount. Whether such an approach is outer aligned or not is likely to depend heavily on the specific metric used to measure the distance from HCH, as well as the robustness of HCH to small deviations according to that metric.

\subsection{Inner alignment} We are using relaxed adversarial training here, which means that---as with all the other relaxed adversarial training approaches---inner alignment is likely to come down to whether or not the overseer is able to produce a consistent gradient away from deception. As always, this approach could perform quite differently from other approaches on this count, since the model helping with oversight and the model being overseen are being trained differently. Comparing this approach to imitative amplification: the addition of an RL objective could help significantly if performance in the RL environment is highly correlated to oversight effectiveness. Furthermore, the fact that the model is being trained to maximize the performance of the human with access to the model means that the model is forced to learn how to present its information to a human to get reward, which could make it significantly more transparent and help with oversight.

On the other hand, explicitly training the model to produce answers which change a human's behavior could also encourage the model to trick the human, which could make deception more likely and thus make oversight harder. Additionally, training on an RL objective could make it significantly harder to do myopia verification, as optimizing even partially for a non-myopic RL loss could act as a strong force pushing the model away from myopia.

\subsection{Training competitiveness} Comparing this approach to imitative amplification: the addition of an RL loss could significantly improve training competitiveness by producing a much richer reward signal, though at the cost of having to run the RL environment. Furthermore, the addition of the imitative loss could even improve training competitiveness relative to pure RL by providing a less sparse reward. Additionally, incorporating an RL loss enables directly training the model to solve long-term tasks involving difficult credit assignment problems that might be hard for imitative amplification alone to handle (though, as in the case of recursive reward modeling, this trades off with the potential safety benefits of myopia).

However, it is unclear what happens to this behavior if the RL loss is annealed away---ideally, if it is possible for HCH to produce the behavior, then we can hope that the model will converge on that, though this requires the optimal RL behavior to be close enough to HCH that first training on the RL loss and then training on the imitative loss actually helps with the imitation task. For that to be the case, success on the RL task likely needs to be highly correlated with good HCH imitation, for which language modeling and human approval maximization tasks (such as the AI safety via debate game, as in ``Synthesizing amplification and debate'') could be good candidates.\cite{synthesizing}

\subsection{Performance competitiveness} Like outer alignment, performance competitiveness is likely to depend heavily on exactly how the different losses are annealed in the limit. As was suggested previously, annealing towards the imitative loss makes the limit HCH, giving this approach the same performance competitiveness properties as imitative amplification. However, again, there are other options, such as using the imitative loss as a constraint such that the model is trained to maximize the RL reward without deviating too far from HCH. Such a constraint could help performance competitiveness compared to training purely on the RL loss (if the resulting model is significantly better at interacting with humans, for example) or hinder performance competitiveness (if it's very difficult to do well on the RL task while staying close to HCH).

\section{Amplification alongside RL + relaxed adversarial training}
\label{sec:11}

In this approach, the goal is to simultaneously train two models: a question-answering system $M$ and an RL agent $\pi$.

We will train $\pi$ on the joint objective of maximizing the RL reward and decreasing the probability of deceptive or otherwise catastrophic behavior as assessed by $\text{Amp}(M)$. Note that while we are using a sort of relaxed adversarial training loss here, the overseer is using a totally different model ($M$) than the model being overseen ($\pi$).

\vspace{4mm}

\begin{figure}[h!]
  \centering
  \includegraphics[width=0.7\textwidth]{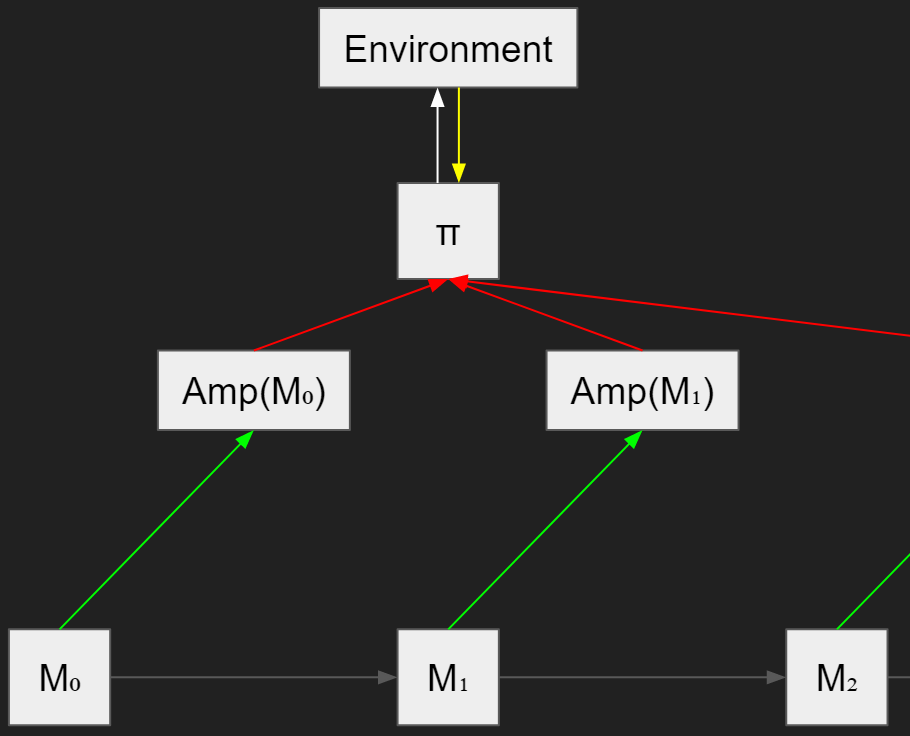}
  \caption{The training process for $\pi$ in amplification alongside RL, where the white arrow indicates $\pi$'s actions in the environment, the yellow arrow indicates $\pi$'s RL loss, and the red arrows indicate the oversight loss for $\pi$.}
\end{figure}

\vspace{2mm}

\noindent We will then train $M$ on the following joint objective. First, we want to include the standard imitative amplification loss and train $M$ to imitate $\text{Amp}(M)$.

\vspace{4mm}

\begin{figure}[H]
  \centering
  \includegraphics[width=0.7\textwidth]{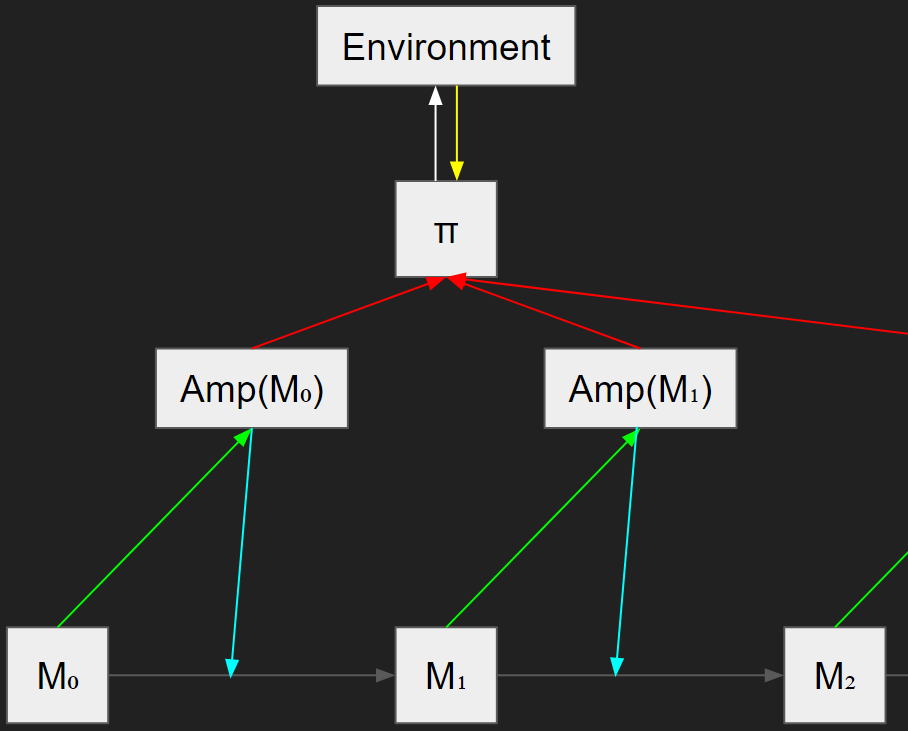}
  \caption{Amplification alongside RL, where $M$ is trained only via an imitative loss.}
\end{figure}

\vspace{2mm}

\noindent Second, we want to train $M$ to cause $\text{Amp}(M)$ to predict $\pi$'s actions and the RL environment's responses well. The idea here is to train $M$ to tell the human everything the human needs to know to understand both what $\pi$ is doing and how to replicate $\pi$'s behavior so that the human with access to the model can match $\pi$'s performance in the RL environment.

\vspace{4mm}

\begin{figure}[H]
  \centering
  \includegraphics[width=0.7\textwidth]{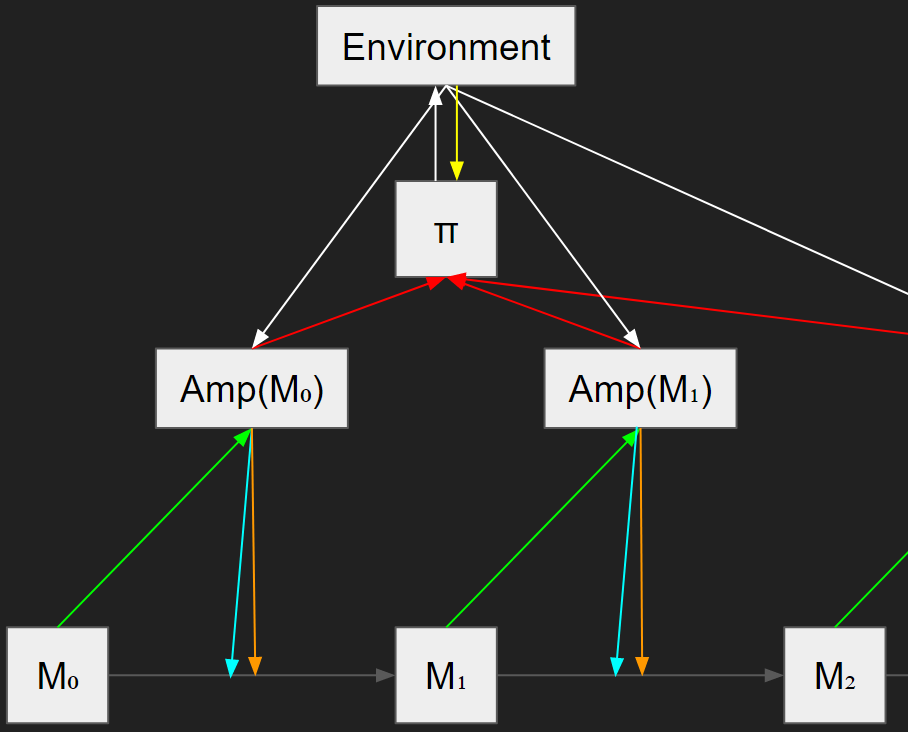}
  \caption{Amplification alongside RL, where the white arrows to $\text{Amp}(M)$ indicate information that $\text{Amp}(M)$ attempts to predict in the RL environment, and the orange arrows indicate the prediction loss for training $M$ to cause $\text{Amp}(M)$ to predict well.}
\end{figure}

\vspace{2mm}

\noindent Third, we can add standard relaxed adversarial training and also train $M$ to decrease the probability of deceptive or otherwise catastrophic behavior as assessed by $\text{Amp}(M)$.

\begin{figure}[H]
    \centering
    \includegraphics[width=0.7\textwidth]{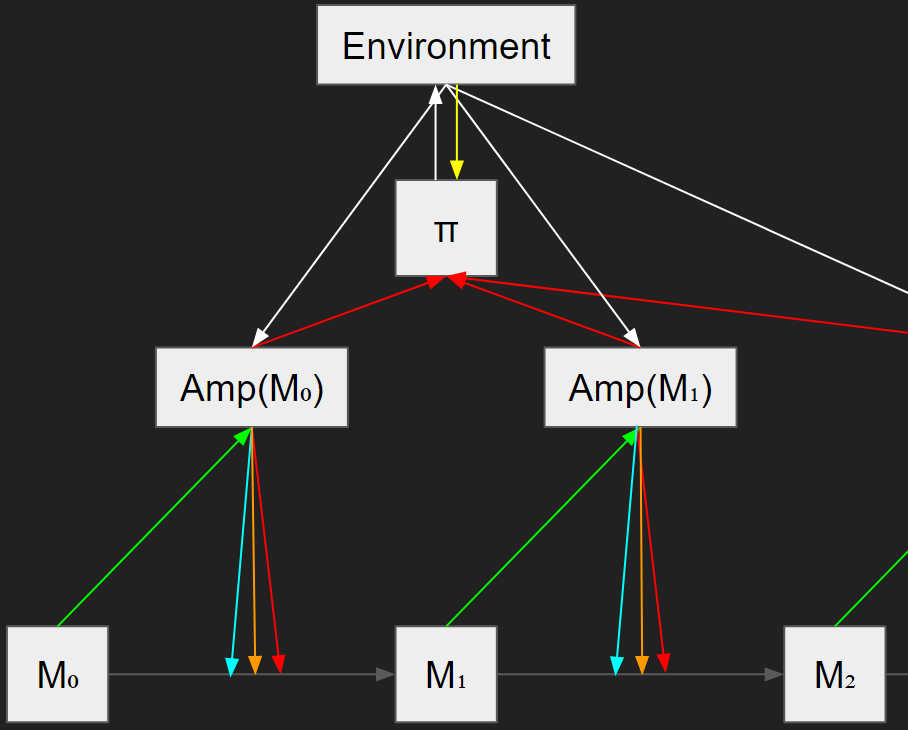}
    \caption{Amplification alongside RL + relaxed adversarial training, where the red arrows indicate the oversight loss.}
\end{figure}

\vspace{2mm}

\noindent Finally, once we have trained both $M$ and $\pi$ via the above losses, a critical component of this approach is that we then do away with $\pi$ completely and exclusively deploy the resulting $M$.

\subsection{Outer alignment} Since we only deploy $M$, not $\pi$, the question of outer alignment here should be entirely dependent on whether this procedure produces outer aligned $M$. Furthermore, similarly to the previous approach, we can anneal the combination of losses here to get the desired limiting behavior and thus control our outer alignment properties. In particular, if we anneal towards the imitative loss, this approach can be made to limit to HCH, giving it the same outer alignment properties as imitative amplification. Even if we don't anneal towards the imitative loss, as long as it is possible for HCH to fully predict both $\pi$ and the RL environment in the limit and the imitative loss isn't annealed away completely, at minimal loss we should still end up with HCH.

\subsection{Inner alignment} Inner alignment is particularly tricky for this approach. First, there's the standard relaxed adversarial training question of whether $\text{Amp}(M)$ will be able to properly oversee $M$. Comparing this approach to the previous approach, the fact that the auxiliary reward is purely predictive rather than a full RL loss should help us here, as it means that our loss is entirely myopic and thus we should be able to do myopia verification. That being said, if $\pi$ is sufficiently non-myopic, then having $M$ predict $\pi$ while still staying myopic itself could be a challenge.

There's also the problem of $\pi$ itself---while we throw away $\pi$ at the end, there are still likely to be opportunities for a sufficiently competent and deceptive $\pi$ to cause serious issues. For example, $\pi$ could choose its actions to influence $M$'s training in a particular way---or just break out of its sandbox and directly take bad actions in the world.

Thus, we need the oversight of $\pi$ by $\text{Amp}(M)$ to at least work sufficiently well to prevent such issues. For that to happen, it needs to be the case that $\text{Amp}(M)$ with access to transparency tools can predict whether (and when) $\pi$ will ever produce such catastrophic behavior. Thus, the fact that we are explicitly training $M$ to cause $\text{Amp}(M)$ to be able to predict $\pi$ could be quite helpful here, as this should hopefully make $\text{Amp}(M)$ good enough at predicting $\pi$ to foresee $\pi$ becoming deceptive or producing any such bad behavior.

\subsection{Training competitiveness} Compared to other amplification approaches, this approach could be significantly more training competitive, as it enables us to just train an RL model with minimal restrictions. However, it also requires us to jointly train both an RL model and an amplification model, which could make it significantly less competitive, especially if it is difficult to successfully train the amplification model to predict the RL model.

Such training can be viewed as effectively just model distillation, which is generally fairly doable even for current models. It should also be noted, however, that the inclusion of the imitative loss will change things, especially if you anneal towards the imitative loss in the limit. In such a situation, the question becomes how useful the auxiliary RL prediction task is in helping with the core imitation task, which---similarly to the previous approach---could be very dependent on the specific RL environment used and how related performance on that RL environment is to the imitation task.

\subsection{Performance competitiveness} As with outer alignment, annealing towards the imitative loss should produce HCH as the limit, giving this approach the same performance competitiveness properties as imitative amplification. Furthermore---again as with outer alignment---even if the imitative loss isn't annealed to completely, as long as HCH can predict $\pi$ in the limit, you should still get HCH at minimal loss.

\bibliography{11_proposals}

\begin{thebibliography}{43}
\providecommand{\natexlab}[1]{#1}
\providecommand{\url}[1]{\texttt{#1}}
\expandafter\ifx\csname urlstyle\endcsname\relax
  \providecommand{\doi}[1]{doi: #1}\else
  \providecommand{\doi}{doi: \begingroup \urlstyle{rm}\Url}\fi

\bibitem[Hubinger(2020{\natexlab{a}})]{post}
Evan Hubinger.
\newblock An overview of 11 proposals for building safe advanced {AI},
  2020{\natexlab{a}}.
\newblock URL
  \url{https://www.alignmentforum.org/posts/fRsjBseRuvRhMPPE5/an-overview-of-11-proposals-for-building-safe-advanced-ai}.

\bibitem[Christiano et~al.(2018)Christiano, Shlegeris, and
  Amodei]{amplification}
Paul Christiano, Buck Shlegeris, and Dario Amodei.
\newblock Supervising strong learners by amplifying weak experts.
\newblock \emph{arXiv}, 2018.
\newblock URL \url{https://arxiv.org/abs/1810.08575}.

\bibitem[Irving et~al.(2018)Irving, Christiano, and Amodei]{debate}
Geoffrey Irving, Paul Christiano, and Dario Amodei.
\newblock {AI} safety via debate.
\newblock \emph{arXiv}, 2018.
\newblock URL \url{https://arxiv.org/abs/1805.00899}.

\bibitem[Leike et~al.(2018)Leike, Krueger, Everitt, Martic, Maini, and
  Legg]{leike}
Jan Leike, David Krueger, Tom Everitt, Miljan Martic, Vishal Maini, and Shane
  Legg.
\newblock Scalable agent alignment via reward modeling: a research direction.
\newblock \emph{arXiv}, 2018.
\newblock URL \url{https://arxiv.org/abs/1811.07871}.

\bibitem[Hubinger et~al.(2019)Hubinger, van Merwijk, Mikulika, Skalse, and
  Garrabrant]{risks}
Evan Hubinger, Chris van Merwijk, Vladimir Mikulika, Joar Skalse, and Scott
  Garrabrant.
\newblock {Risks from Learned Optimization in Advanced Machine Learning
  Systems}.
\newblock \emph{arXiv}, 2019.
\newblock URL \url{https://arxiv.org/abs/1906.01820}.

\bibitem[Hubinger(2020{\natexlab{b}})]{market_making}
Evan Hubinger.
\newblock {AI} safety via market making, 2020{\natexlab{b}}.
\newblock URL
  \url{https://www.alignmentforum.org/posts/YWwzccGbcHMJMpT45/ai-safety-via-market-making}.

\bibitem[Bostrom(2014)]{superintelligence}
Nick Bostrom.
\newblock \emph{Superintelligence: Paths, Dangers, Strategies}.
\newblock Oxford University Press, 2014.
\newblock URL
  \url{https://global.oup.com/academic/product/superintelligence-9780199678112?cc=us&lang=en&}.

\bibitem[Hubinger(2020{\natexlab{c}})]{outer_alignment}
Evan Hubinger.
\newblock Outer alignment and imitative amplification, 2020{\natexlab{c}}.
\newblock URL
  \url{https://www.alignmentforum.org/posts/33EKjmAdKFn3pbKPJ/outer-alignment-and-imitative-amplification}.

\bibitem[Soares et~al.(2015)Soares, Fallenstein, Yudkowsky, and
  Armstrong]{corrigibility}
Nate Soares, Benja Fallenstein, Eliezer Yudkowsky, and Stuart Armstrong.
\newblock Corrigibility.
\newblock \emph{AAAI 2015}, 2015.
\newblock URL \url{https://intelligence.org/files/Corrigibility.pdf}.

\bibitem[Hubinger(2019{\natexlab{a}})]{mechanistic}
Evan Hubinger.
\newblock Towards a mechanistic understanding of corrigibility,
  2019{\natexlab{a}}.
\newblock URL
  \url{https://www.alignmentforum.org/posts/BKM8uQS6QdJPZLqCr/towards-a-mechanistic-understanding-of-corrigibility}.

\bibitem[Baker et~al.(2019)Baker, Kanitscheider, Markov, Wu, Powell, McGrew,
  and Mordatch]{tool_use}
Bowen Baker, Ingmar Kanitscheider, Todor Markov, Yi~Wu, Glenn Powell, Bob
  McGrew, and Igor Mordatch.
\newblock {Emergent Tool Use From Multi-Agent Autocurricula}.
\newblock \emph{arXiv}, 2019.
\newblock URL \url{https://arxiv.org/abs/1909.07528}.

\bibitem[Ngo(2020)]{multi_agent_safety}
Richard Ngo.
\newblock Multi-agent safety, 2020.
\newblock URL
  \url{https://www.alignmentforum.org/posts/BXMCgpktdiawT3K5v/multi-agent-safety}.

\bibitem[Olah et~al.(2020)Olah, Cammarata, Schubert, Goh, Petrov, and
  Carter]{circuits}
Chris Olah, Nick Cammarata, Ludwig Schubert, Gabriel Goh, Michael Petrov, and
  Shan Carter.
\newblock {Thread: Circuits}.
\newblock \emph{Distill}, 2020.
\newblock URL \url{https://distill.pub/2020/circuits/}.

\bibitem[Christiano(2016{\natexlab{a}})]{catastrophes}
Paul Christiano.
\newblock Learning with catastrophes, 2016{\natexlab{a}}.
\newblock URL
  \url{https://ai-alignment.com/learning-with-catastrophes-59387b55cc30}.

\bibitem[Hubinger(2019{\natexlab{b}})]{chris_olah}
Evan Hubinger.
\newblock {Chris Olah’s views on AGI safety}, 2019{\natexlab{b}}.
\newblock URL
  \url{https://www.alignmentforum.org/posts/X2i9dQQK3gETCyqh2/chris-olah-s-views-on-agi-safety}.

\bibitem[Hubinger(2019{\natexlab{c}})]{adversarial_ida}
Evan Hubinger.
\newblock {A Concrete Proposal for Adversarial IDA}, 2019{\natexlab{c}}.
\newblock URL
  \url{https://www.alignmentforum.org/posts/jYvm4mmjvGHcPXtGL/a-concrete-proposal-for-adversarial-ida}.

\bibitem[Christiano(2016{\natexlab{b}})]{strong_hch}
Paul Christiano.
\newblock Strong {HCH}, 2016{\natexlab{b}}.
\newblock URL \url{https://ai-alignment.com/strong-hch-bedb0dc08d4e}.

\bibitem[Christiano(2019)]{universality}
Paul Christiano.
\newblock Universality and consequentialism within {HCH}, 2019.
\newblock URL
  \url{https://ai-alignment.com/universality-and-consequentialism-within-hch-c0bee00365bd}.

\bibitem[Christiano(2015)]{efficient_feedback}
Paul Christiano.
\newblock Efficient feedback, 2015.
\newblock URL \url{https://ai-alignment.com/efficient-feedback-a347748b1557}.

\bibitem[Hubinger(2019{\natexlab{d}})]{relaxed}
Evan Hubinger.
\newblock Relaxed adversarial training for inner alignment, 2019{\natexlab{d}}.
\newblock URL
  \url{https://www.alignmentforum.org/posts/9Dy5YRaoCxH9zuJqa/relaxed-adversarial-training-for-inner-alignment}.

\bibitem[Hubinger(2019{\natexlab{e}})]{gradient_hacking}
Evan Hubinger.
\newblock Gradient hacking, 2019{\natexlab{e}}.
\newblock URL
  \url{https://www.alignmentforum.org/posts/uXH4r6MmKPedk8rMA/gradient-hacking}.

\bibitem[Warnell et~al.(2017)Warnell, Waytowich, Lawhern, and
  Stone]{deep_tamer}
Garrett Warnell, Nicholas Waytowich, Vernon Lawhern, and Peter Stone.
\newblock {Deep TAMER: Interactive Agent Shaping in High-Dimensional State
  Spaces}.
\newblock \emph{arXiv}, 2017.
\newblock URL \url{https://arxiv.org/abs/1709.10163}.

\bibitem[Arumugam et~al.(2019)Arumugam, Lee, Saskin, and Littman]{deep_rl}
Dilip Arumugam, Jun~Ki Lee, Sophie Saskin, and Michael~L. Littman.
\newblock {Deep Reinforcement Learning from Policy-Dependent Human Feedback}.
\newblock \emph{arXiv}, 2019.
\newblock URL \url{https://arxiv.org/abs/1902.04257}.

\bibitem[Christiano(2014)]{model_free}
Paul Christiano.
\newblock Approval-directed agents, 2014.
\newblock URL \url{https://ai-alignment.com/model-free-decisions-6e6609f5d99e}.

\bibitem[Olah(2015)]{visualizing}
Chris Olah.
\newblock {Visualizing Representations: Deep Learning and Human Beings}, 2015.
\newblock URL
  \url{https://colah.github.io/posts/2015-01-Visualizing-Representations/}.

\bibitem[Christiano(2016{\natexlab{c}})]{universal_prior}
Paul Christiano.
\newblock What does the universal prior actually look like?,
  2016{\natexlab{c}}.
\newblock URL
  \url{https://ordinaryideas.wordpress.com/2016/11/30/what-does-the-universal-prior-actually-look-like}.

\bibitem[Demski(2019)]{partial_agency}
Abram Demski.
\newblock {Partial Agency}, 2019.
\newblock URL \url{https://www.alignmentforum.org/s/HeYtBkNbEe7wpjc6X}.

\bibitem[Radford et~al.(2019)Radford, Wu, Child, Luan, Amodei, and
  Sutskever]{language_models}
Alec Radford, Jeffrey Wu, Rewon Child, David Luan, Dario Amodei, and Ilya
  Sutskever.
\newblock {Language Models are Unsupervised Multitask Learners}.
\newblock \emph{{OpenAI}}, 2019.
\newblock URL
  \url{https://cdn.openai.com/better-language-models/language_models_are_unsupervised_multitask_learners.pdf}.

\bibitem[Kumar and Garrabrant(2019)]{human_models}
Ramana Kumar and Scott Garrabrant.
\newblock Thoughts on human models.
\newblock \emph{MIRI}, 2019.
\newblock URL
  \url{https://intelligence.org/2019/02/22/thoughts-on-human-models}.

\bibitem[Kusumoto et~al.(2018)Kusumoto, Yahata, and Sakai]{theorem_proving}
Mitsuru Kusumoto, Keisuke Yahata, and Masahiro Sakai.
\newblock {Automated Theorem Proving in Intuitionistic Propositional Logic by
  Deep Reinforcement Learning}.
\newblock \emph{arXiv}, 2018.
\newblock URL \url{https://arxiv.org/abs/1811.00796}.

\bibitem[Bansal et~al.(2019)Bansal, Loos, Rabe, Szegedy, and Wilcox]{holist}
Kshitij Bansal, Sarah~M. Loos, Markus~N. Rabe, Christian Szegedy, and Stewart
  Wilcox.
\newblock {HOList: An Environment for Machine Learning of Higher-Order Theorem
  Proving}.
\newblock \emph{arXiv}, 2019.
\newblock URL \url{https://arxiv.org/abs/1904.03241}.

\bibitem[Senior et~al.(2020)Senior, Evans, Jumper, Kirkpatrick, Sifre, Green,
  Qin, Žídek, Nelson, Bridgland, Penedones, Petersen, Simonyan, Crossan,
  Kohli, Jones, Silver, Kavukcuoglu, and Hassabis]{protein}
Andrew~W. Senior, Richard Evans, John Jumper, James Kirkpatrick, Laurent Sifre,
  Tim Green, Chongli Qin, Augustin Žídek, Alexander W.~R. Nelson, Alex
  Bridgland, Hugo Penedones, Stig Petersen, Karen Simonyan, Steve Crossan,
  Pushmeet Kohli, David~T. Jones, David Silver, Koray Kavukcuoglu, and Demis
  Hassabis.
\newblock Improved protein structure prediction using potentials from deep
  learning.
\newblock \emph{Nature}, 2020.
\newblock URL \url{https://www.nature.com/articles/s41586-019-1923-7.epdf}.

\bibitem[Bostrom(2019)]{vulnerable}
Nick Bostrom.
\newblock {The Vulnerable World Hypothesis}.
\newblock \emph{{Global Policy}}, 2019.
\newblock URL \url{https://nickbostrom.com/papers/vulnerable.pdf}.

\bibitem[Sandberg and Bostrom(2008)]{wbe}
Anders Sandberg and Nick Bostrom.
\newblock {Whole Brain Emulation: A Roadmap}.
\newblock \emph{{FHI}}, 2008.
\newblock URL
  \url{https://www.fhi.ox.ac.uk/brain-emulation-roadmap-report.pdf}.

\bibitem[Ren et~al.(2019)Ren, Liu, Fertig, Snoek, Poplin, DePristo, Dillon, and
  Lakshminarayanan]{out_of_distribution}
Jie Ren, Peter~J. Liu, Emily Fertig, Jasper Snoek, Ryan Poplin, Mark~A.
  DePristo, Joshua~V. Dillon, and Balaji Lakshminarayanan.
\newblock {Likelihood Ratios for Out-of-Distribution Detection}.
\newblock \emph{arXiv}, 2019.
\newblock URL \url{https://arxiv.org/abs/1906.02845}.

\bibitem[Ho and Ermon(2016)]{generative}
Jonathan Ho and Stefano Ermon.
\newblock {Generative Adversarial Imitation Learning}.
\newblock \emph{arXiv}, 2016.
\newblock URL \url{https://arxiv.org/abs/1606.03476}.

\bibitem[Fu et~al.(2017)Fu, Luo, and Levine]{learning_robust_rewards}
Justin Fu, Katie Luo, and Sergey Levine.
\newblock {Learning Robust Rewards with Adversarial Inverse Reinforcement
  Learning}.
\newblock \emph{arXiv}, 2017.
\newblock URL \url{https://arxiv.org/abs/1710.11248}.

\bibitem[Drexler(2019)]{reframing_si}
K.~Eric Drexler.
\newblock {Reframing Superintelligence: Comprehensive AI Services as General
  Intelligence}.
\newblock \emph{{FHI}}, 2019.
\newblock URL
  \url{https://www.fhi.ox.ac.uk/wp-content/uploads/Reframing_Superintelligence_FHI-TR-2019-1.1-1.pdf}.

\bibitem[Barnes and Christiano(2020)]{debate_progress}
Beth Barnes and Paul Christiano.
\newblock {Writeup: Progress on AI Safety via Debate}, 2020.
\newblock URL
  \url{https://www.alignmentforum.org/posts/Br4xDbYu4Frwrb64a/writeup-progress-on-ai-safety-via-debate-1}.

\bibitem[Silver et~al.(2018)Silver, Hubert, Schrittwieser, Antonoglou, Lai,
  Guez, Lanctot, Sifre, Kumaran, Graepel, Lillicrap, Simonyan, and
  Hassabis]{go}
David Silver, Thomas Hubert, Julian Schrittwieser, Ioannis Antonoglou, Matthew
  Lai, Arthur Guez, Marc Lanctot, Laurent Sifre, Dharshan Kumaran, Thore
  Graepel, Timothy Lillicrap, Karen Simonyan, and Demis Hassabis.
\newblock A general reinforcement learning algorithm that masters chess, shogi,
  and {Go} through self-play.
\newblock \emph{Science}, 2018.
\newblock URL
  \url{https://science.sciencemag.org/content/362/6419/1140.full?ijkey=XGd77kI6W4rSc&keytype=ref&siteid=sci}.

\bibitem[Wolski et~al.(2018)Wolski, Sidor, Petrov, Farhi, Raiman, Zhang,
  Brockman, Dennison, Tang, Pondé, Chan, Pachocki, and Dębiak]{openai_five}
Filip Wolski, Szymon Sidor, Michael Petrov, David Farhi, Jonathan Raiman, Susan
  Zhang, Greg Brockman, Christy Dennison, Jie Tang, Henrique Pondé, Brooke
  Chan, Jakub Pachocki, and Przemysław Dębiak.
\newblock {OpenAI Five}, 2018.
\newblock URL \url{https://openai.com/blog/openai-five/}.

\bibitem[{The AlphaStar team}(2019)]{alphastar}
{The AlphaStar team}.
\newblock {AlphaStar: Mastering the Real-Time Strategy Game StarCraft II},
  2019.
\newblock URL
  \url{https://deepmind.com/blog/article/alphastar-mastering-real-time-strategy-game-starcraft-ii}.

\bibitem[Hubinger(2020{\natexlab{d}})]{synthesizing}
Evan Hubinger.
\newblock Synthesizing amplification and debate, 2020{\natexlab{d}}.
\newblock URL
  \url{https://www.alignmentforum.org/posts/dJSD5RK6Qoidb3QY5/synthesizing-amplification-and-debate}.

\end{thebibliography}

\end{document}